\documentclass[10pt,leqno]{amsart}
\usepackage{graphicx}
\usepackage[utf8]{inputenc}
\usepackage{indentfirst,csquotes}

\usepackage{amssymb,amsthm,amsmath}
\usepackage[dvipsnames]{xcolor}
\usepackage{paralist,fancyhdr,etoolbox}

\pdfoutput=1
\usepackage[utf8]{inputenc}
\usepackage[T1]{fontenc}
\usepackage{booktabs,microtype,caption,subcaption}
\usepackage[super,numbers]{natbib}
\usepackage{bibunits}
\usepackage{hyperref}
\hypersetup{colorlinks=true,linkcolor=black,citecolor=black,filecolor=black,urlcolor=black}
\renewcommand{\cite}{\citep}
\definecolor{cadmiumgreen}{rgb}{0.0,0.42,0.24}
\definecolor{cardinal}{rgb}{0.77,0.12,0.23}
\providecommand{\sym}[1]{\ifmmode^{#1}\else\(^{#1}\)\fi}

\makeatletter
\AtBeginDocument{%
\patchcmd{\NAT@hyper@}{\@citeb}{\@bibunitname.\@citeb}{}{\errmessage{Citation link patch failed}}
\let\originalnatanchorstart\hyper@natanchorstart
\def\hyper@natanchorstart#1{\originalnatanchorstart{\@bibunitname.#1}}
\let\originalnatlinkbreak\hyper@natlinkbreak
\def\hyper@natlinkbreak#1#2{\originalnatlinkbreak{#1}{\@bibunitname.#2}}
}
\makeatother
\begin{document}
\baselineskip=16pt

\begin{bibunit}

\newcommand{\logo}{%
    \raisebox{-0.2\height}{\includegraphics[scale=0.3]{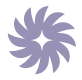}}%
}

\title[Tutor CoPilot: A Human-AI Approach]{\logo{} Tutor CoPilot: A Human-AI Approach for \\ Scaling Real-Time Expertise}
\author[Wang, Ribeiro, Robinson, Loeb, and Demszky]{Rose E. Wang, Ana T. Ribeiro, Carly D. Robinson, Susanna Loeb, and Dora Demszky}

\date{}

\newcommand{\pencil}{
    \includegraphics[scale=0.09]{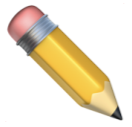}}

\newcommand{\refresh}{
    \includegraphics[scale=0.09]{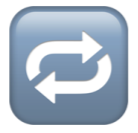}}
    
\newcommand{\dropdown}{
    \includegraphics[scale=0.09]{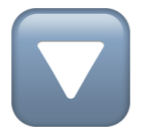}}

\thanks{Stanford University, Stanford, CA 94304. Rose E. Wang and Ana T. Ribeiro are co-first authors, having contributed equally to the project. The project is also an equal collaboration of Loeb's and Demszky's research labs. Our pre-registration for this randomized controlled trial can be found here: \url{https://osf.io/8d6ha}. This study was approved under Stanford University’s IRB Protocol ``National Student Support Accelerator: Effects of tutoring at school district test sites'' (\#68027). We have complied with all relevant ethical regulations for studies involving human research participants. Correspondence should be addressed to A.T.R.: anactr@stanford.edu. }

\begin{abstract}

\noindent Generative AI could expand access to expert guidance in education, where such support is scarce. We introduce Tutor CoPilot, a human-AI system that models expert thinking to assist tutors in real time. In a randomized controlled trial with more than 700 tutors and 1,000 students from underserved communities, students whose tutors used Tutor CoPilot were 4 percentage points more likely to master mathematics topics (p<.01). Gains were largest for students of lower-rated tutors (9 percentage points), while annual costs were about \$20 per tutor. Analysis of more than 350,000 messages indicates that Tutor CoPilot promotes effective pedagogy, increasing probing questions and reducing generic praise. These findings demonstrate the potential of human-AI systems to scale expertise and improve educational quality.

\end{abstract}

\maketitle

\section*{Introduction}

Generative AI, including Large Language Models (LLMs), has the potential to transform real-world domains like education, healthcare and law, which rely on a skilled workforce to handle complex tasks. 
For example, educators (e.g., teachers and tutors) are often trained to synthesize curriculum knowledge and recognize student needs in order to provide high-quality learning experiences~\cite{shulman1986those,lampert2001teaching, willingham2008critical}. Traditionally, experts train novices by sharing their intuition and reasoning refined over years of practice~\citep{darling2017effective}.
However, expert-guided training is costly and difficult to scale ~\citep{KraftBlazarHogan2018, kelly2020using}. Knight and Skrtic (2021)\citep{knight2021cost} estimate an annual cost of at least US\$4,800 per teacher for a coaching program, about 2-4\% of district expenditures in the US, with national investments reaching tens of billions of dollars.

Traditional training programs not only demand significant time outside of instructional hours, something many part-time educators cannot manage~\citep{yoon2007reviewing},
but they also tend to follow static curricula that are often misaligned to the practical needs of novice educators~\citep{van2012makes,boyd2009teacher}. 
As a result, many novice educators do not have access to expert guidance and must develop their skills on the job, resulting in students likely missing out on valuable learning opportunities, which can be particularly harmful for students who have the most to gain from improved educational experiences~\citep{darling2006highly,boyd2005explaining}.

LLMs may be able to provide real-time guidance for novices at scale, but several challenges must be addressed to make this feasible.
LLMs are trained on Web data (e.g., Wikipedia and Reddit) which differ substantially from real-world K-12 interactions, thus out-of-the-box LLMs often fail in real-world learning settings~\citep{wang2023a}. 
Current techniques to adapt LLMs for real-world settings (e.g., fine-tuning or prompt-engineering) struggle to elicit appropriate behaviors because these approaches focus on surface-level language patterns and overlook the latent reasoning processes that expert educators have honed through years of practice to guide their decision-making~\citep{jurenka2024towards, Singer_2023, wang-etal-2024-bridging}. 
LLMs also lack real-world knowledge that is important for delivering high-quality learning experiences, such as knowledge of the curriculum, previous interactions, and future learning objectives. 
In contrast, human educators possess this contextual knowledge \citep{wang-etal-2024-bridging}. 
Their knowledge may complement traditional LLMs to produce more effective approaches to supporting novices at scale.

\begin{figure}[!t]
    \centering
    \includegraphics[width=\linewidth]{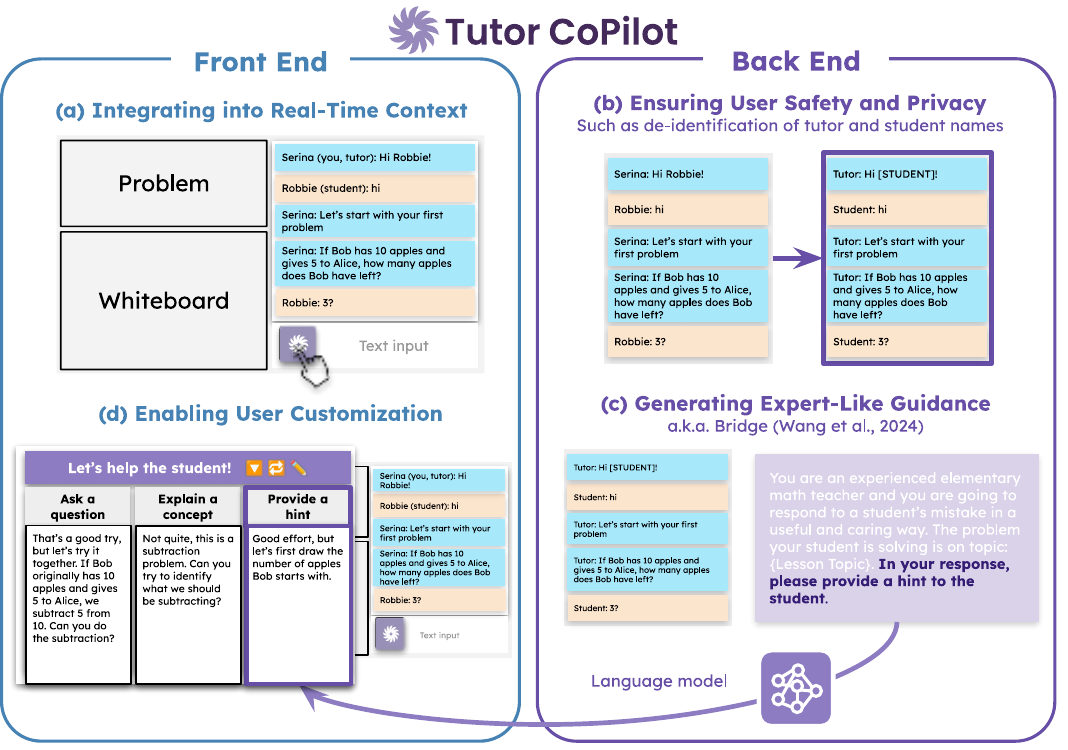}
    \caption{\textbf{Illustration of Tutor CoPilot front end interface and back end system operations.}
    Panel (a) \textit{Tutor CoPilot is integrated into live contexts} as a button that the tutor can activate for real-time assistance during their tutoring sessions.
    Panel (b) \textit{Tutor CoPilot applies user safety and privacy practices}, such as automatically de-identifying student and tutor names and limiting the amount of user information sent to external LM services.
    Panel (c) \textit{Tutor CoPilot generates expert-like guidance} by leveraging the Bridge method~\citep{wang-etal-2024-bridging} which captures expert decision-making from their verbalized reasoning patterns.
    Panel (d) \textit{Tutor CoPilot is activated}, displaying three suggested responses for tutors to choose from in panels that 
    enable user customization by editing (\pencil), re-generating (\refresh), or displaying the full list of strategies with seven additional options (\dropdown).}
    \label{fig:tutor_copilot}
\end{figure}

We introduce Tutor CoPilot, a Human-AI approach to scale expertise by providing real-time suggestions to tutors remediating students' mathematical mistakes.
Our work builds on prior research by Wang et al. (2024a)\cite{wang-etal-2024-bridging}, which used think-aloud protocols to capture experienced educators' reasoning and adapted LLMs to generate expert-like suggestions.
Tutor CoPilot aims to enhance the quality of K-12 education at scale by delivering actionable guidance tutors can immediately apply during live sessions. The tool is tutor-facing only and offers multiple pedagogical strategies whenever activated, preserving tutor agency. Tutors retain control over whether and how to use these suggestions, allowing them to leverage their contextual knowledge. This approach can improve the quality and effectiveness of tutors’ responses while also supporting their professional growth \citep{smith2017teachers,brod2023agency,calvert2016power}.
Figure~\ref{fig:tutor_copilot} illustrates the user interface for tutors in panels (a) and (d), and the backend process for generating suggestions once the tool is activated in panels (b) and (c). Supplementary Information section~\ref{app:tutor_copilot} includes further details.

This preregistered study presents the first randomized controlled trial of a Human-AI system in live tutoring.
In collaboration with FEV Tutor, a chat-based virtual tutoring provider, and a U.S. Southern school district, we conducted an intervention through an in-school, virtual tutoring program for mathematics with more than 700 tutors and 1000 K-12 students from schools that receive federal funding to support students from low-income families.   
This study aims to answer four research questions (RQs): 

\begin{enumerate}
    \item To what extent does Tutor CoPilot affect student learning? 
    \item Does the effect of Tutor CoPilot on student learning differ by the initial effectiveness of tutors? 
    \item How does Tutor CoPilot change tutoring quality as measured by the language tutors use with students? 
    \item How do tutors perceive Tutor CoPilot? 
\end{enumerate}

Our findings reveal that Tutor CoPilot significantly improves student learning outcomes (RQ1): Our intent-to-treat analysis shows students whose tutors have access to Tutor CoPilot are 4 percentage points (p.p.) more likely to master session lesson topics. 
Tutor CoPilot particularly benefits lower-rated and less-experienced tutors (RQ2), with these tutors improving their students' mastery by up to 9 percentage points over the control, according to a heterogeneity analysis based on tutor ratings from the tutoring provider. 
Tutors with access to Tutor CoPilot are more likely to use high-quality strategies that foster student understanding (RQ3), determined by classifiers that identify high- and low-quality pedagogical strategies.  
Although our study design and implementation were not set up to estimate the direct impact of Tutor CoPilot on end-of-the-year (EOY) student achievement scores on NWEA MAP, a back-of-the-envelope calculation suggests that a full year of tutoring with Tutor CoPilot enabled may improve student performance on the test by 0.024 standard deviations (Supplementary Information section~\ref{app:exit_ticket}).
Finally, in our interviews with tutors, tutors reported that Tutor CoPilot was helpful but indicated room for improvement in its guidance, such as by generating appropriate grade-level language. With an estimated annual cost of just \$20 per tutor based on usage patterns (Supplementary Information section~\ref{app:study_cost}), Tutor CoPilot offers a scalable and cost-effective alternative to traditional, resource-intensive training programs. Overall, our findings demonstrate that Tutor CoPilot is a promising Human-AI approach combining LLMs with task-specific expertise, enhancing educational quality for students from underserved communities.

\section*{Methods}
The implementation part of our study lasted for about 7 weeks, starting with the launch of Tutor CoPilot for treatment tutors on the platform on March 27th, and ending on May 17th, 2024, with the end of the school year.
We preregistered our primary hypotheses and analysis plan on the Open Science Framework platform prior to accessing the data, which can be found: \url{https://osf.io/8d6ha}.  

\subsection*{Randomization and Participants}
\vspace{-.5em}

We partnered with FEV Tutor, a chat-based tutoring provider, and a large southern U.S. school district for this study.
Students receiving tutoring through FEV Tutors were matched to an available tutor upon their logging into the platform but were not guaranteed to consistently match with the same tutor. Considering that Tutor CoPilot is a tutor-facing tool only that requires training, we opted for randomizing tutors into the treatment (access to Tutor CoPilot) or control (business as usual) conditions. That is, students were not randomized into one of the two conditions, and their treatment status could vary from session to session, depending on the treatment assignment of the tutor with whom they were matched. For this reason, we are not able to estimate the impact of Tutor CoPilot directly on student EOY achievement scores, but rather focus our analysis on session-level outcomes for students, such as session exit tickets. 

\paragraph{Tutors.} FEV Tutors initially identified 900 tutors who were assigned to work with students from our study's school district and randomly assigned to the treatment (450) or control (450) conditions. However, due to attrition between the initial assignment time and the study launch, only 872 responded to our pre-launch survey, and of those, 783 (treatment = 388, control = 395)  had tutoring sessions with a student from our partner district for this study after Tutor CoPilot was launched. Balance checks show that the control and treatment groups do not differ statistically on any baseline variables; these results are detailed in Supplementary Information section~\ref{app:balance}.

The tutors were paid, full-time professionals employed by FEV Tutor and had approximately two years of working experience with the tutoring company on average (Table~\ref{tab:balance_prestudy}). All tutors were required to participate in professional development sessions, which included proprietary training modules on math remediation strategies, independent of the Tutor CoPilot tool. This training promoted similar practices of responding to student mistakes and not immediately giving away answers, and provided examples of student mistakes and example responses to be used. Tutors assigned to the treatment condition received additional training on how to use the Tutor CoPilot tool prior to its launch to minimize the novelty effect. Tutor CoPilot training for treatment tutors is detailed in Supplementary Information section~\ref{app:tutor_copilot_training}. While, in theory, the additional professional development for tutors receiving access to Tutor CoPilot may have affected their practice and student learning directly, any effect is likely to have been small given that all tutors were receiving similar information in their initial training.

\paragraph{Students.} The school district serves more than 30,000 students, with the majority from ethnic/racial minority groups and economically disadvantaged backgrounds. A total of 1,534 students from nine schools in this school district were initially identified as eligible to participate in our study by the district. Student eligibility to participate in our study was determined by student eligibility for tutoring services according to the state policy for administering accelerated services, which meant performing below grade level on the state test the previous spring.  We restrict the sample of students in our study analyses to 1,013 students who attended at least one tutoring session with a tutor enrolled in the study and randomly assigned to the treatment or control condition.
The study was conducted with elementary and middle school students in grades 3-6.
The majority (79\%) of our sample identifies as Hispanic, and 67\% of the sample is classified as economically disadvantaged. 
The full sample description is provided in Supplementary Information section~\ref{app:descriptive}.
\vspace{-1em}
\subsection*{Data}
\vspace{-1em}
Our study uses three types of data about the tutor, student and session. 
Supplementary Information section~\ref{app:study_data} provides more details on our study's data.

For tutors, we collected their treatment group assignment, gender, pre-study quality rating, and pre-study tutoring experience.
The quality rating is the provider's internal measure of quality and is based on manual observations of a random sample of the tutor's sessions, averaged from scores on the provider's proprietary observation rubric; it takes on continuous values between -1 and 1. 
The tutor's tutoring experience is the number of months the tutor had been working with the tutoring provider leading up to the start of the study. 

For students, we obtained data about their gender, race, pre-/post-study NWEA MAP Math and Reading Scores, and other relevant covariates. 
The NWEA MAP assessments are standardized tests administered three times a year, tracking students' academic growth over time. 

We tracked session outcomes such as the number of participation points the tutor awarded the student for their engagement and effort during the tutoring session, whether the student attempted and passed the lesson exit ticket, and student post-session survey responses.
Exit tickets are particularly critical as they assess student mastery of each topic and directly influence the student’s progression through the curriculum. 
Students must pass the exit ticket to advance to the next lesson, meaning students who struggle on exit tickets may encounter delays in their learning progression, limiting their exposure to new material. 
As a result, the exit ticket pass rate is not just a measure of immediate comprehension—it plays a key role in determining the pace at which students advance through their academic content.
Moreover, the number of completed exit tickets by a student is a significant predictor of future MAP test performance -- one additional exit ticket completed is associated with a 0.02 standard deviation increase on the test; we report this in Supplementary Information section~\ref{app:exit_ticket}.

Additionally, we collected each session's chat activity and Tutor CoPilot usage. 
Our final analysis sample includes 4,136 sessions, 2,014 with treatment tutors and 2,122 with control tutors. This translates into 350,000+ chat messages exchanged between tutors and students, more than 170,000 from treatment sessions and 184,000 from control sessions. 
We observe 2,000+ uses of Tutor CoPilot, defined as the number of clicks on the tool during the session that correspond to either an initial activation of Tutor CoPilot (Figure~\ref{fig:tutor_copilot} a) or the tutor clicking on a different strategy from the drop-down options (Figure~\ref{fig:tutor_copilot} d). We do not have data distinguishing different types of clicks.

\subsection*{Analysis}
This study tested the causal impact of Tutor CoPilot on student learning outcomes and tutor practices. 
Below, we summarize our preferred methods for analyzing the following research questions. Following our pre-registration, we imputed missing covariates for students and tutors. For categorical variables, we added an additional ``missing'' category. For continuous variables, we assigned the predicted value based on the other present covariates.  Our initial preregistered outcome was EOY MAP Math achievement scores. The implementation resulted in less consistent treatment exposure than expected, however, so we focus on the available session-level outcomes for assessing the impact of tutor copilot on academic outcomes. See Supplementary Information section \ref{app:student_level} for information on the distribution of treatment exposure and the student-level analysis.

\paragraph{RQ1: Impact on Student Outcomes.} We employed an intent-to-treat (ITT) regression analysis to determine how offering the Tutor CoPilot to tutors predicts student outcomes. 
Equation~\ref{eq:primary} describes our primary model:

\begin{align}
    Y_{ist} = \beta_0 + \beta_1 \text{Treatment}_{t} + X_s\gamma  + \omega_{k(s)} + \epsilon_{ist}
    \label{eq:primary}
\end{align}

where $Y_{ist}$ is the session $i$ outcome of interest (e.g., exit ticket passed) for student $s$ working with tutor $t$ in a given class (combination of school and grade) $k$, $\text{Treatment}_{t}$ is the indicator for whether the session tutor $t$ is in the treatment group, $X_s$ is a vector of student-level covariates, $\omega_{k(s)}$ is the fixed effect for class (school and grade), and $\epsilon_{ist}$ is the residual clustered at the student-tutor pair level to account for correlated observations when students are matched to the same tutor more than once.
The student covariates include categorical indicators for the student's gender, race, free and reduced lunch, special education, and limited English proficiency, as well as a continuous variable of the student's pre-study MAP math score. 
To account for the multiple session-level outcomes we examine, we also report Romano-Wolf adjusted p-values for our treatment estimates \citep{clarke2021romano}. See Supplementary Information section ~\ref{app:alternative_specs} for alternative model specifications.

While the ITT analysis captures the overall impact of the intervention, it ignores whether the tutor used Tutor CoPilot during a session. 
To disentangle the effect of merely having access to the tool from the effect of actually using it, we extended our analysis using a two-stage least squares (2SLS) regression and report those results in the Supplementary Information section~\ref{app:tot}.

\paragraph{RQ2: Heterogeneity by Tutor Initial Effectiveness.} We performed a heterogeneity analysis using the exit ticket passing rate as the primary outcome from our primary model (Equation~\ref{eq:primary}).
Specifically, we categorized the tutor's quality rating and tutoring experience into tercile indicators and interacted these indicators with the treatment indicator to examine how the effect of Tutor CoPilot on students' exit ticket passing rate varies based on the tutor's initial effectiveness. 
This approach allowed us to assess whether more or less effective tutors benefit differently from using Tutor CoPilot in terms of improving student performance.

\paragraph{RQ3: Impact on Tutoring Strategies.}We first defined the taxonomy of strategies based on Wang et al.~(2024a)\citep{wang-etal-2024-bridging} and salient response patterns in our data. We distinguish between higher-quality strategies that seek to probe and scaffold students' problem-solving from lower-quality strategies that lead students directly to the solution or provide generic feedback~\citep{wang-etal-2024-bridging, chi1989selfexplanation, lepper2002wisdom, easley1975teaching, carpenter2003thinking, carpenter1999children, loewenberg2009work, collins2018cognitive}. Although it is worth noting that the quality of strategies may vary by context, such as tutors sometimes giving away answers in pedagogically valuable ways (e.g., by prompting the student afterward to self-explain why that answer was correct), we do not observe tutors giving away answers in such pedagogically meaningful ways in our tutoring context, as they tend to employ scripted, pedagogically poor techniques \citep{wang-etal-2024-bridging}.

Second, we trained machine learning classifiers to identify these strategies at scale.
We did so by (i) randomly sampling 3,000 tutor messages, blind to study condition, (ii) prompting GPT-4 with our taxonomy of strategies to perform a first pass of labeling these examples, (iii) manually verifying and fixing labels, (iv) fine-tuning RoBERTa models on the labeled data with a class-balancing loss~\citep{liu2019roberta,cui2019class}.
Our classifiers achieved high performance, with F1-scores ranging between 0.65 and 0.90. Supplementary Information section~\ref{app:nlp_classifiers} provides more details on this setup.

Finally, we used these classifiers to identify the strategies in our entire data of 350,000+ messages. We then estimated the impact of Tutor CoPilot access on the likelihood of utilizing each strategy by analyzing classifications of messages from treatment relative to control tutors. We used the following logit regression, where $C_{mst}$ is a binary indicator equal to 1 if message $m$, sent by tutor $t$ to student $s$, was identified as the strategy of interest, and $\text{Treatment}_{t}$ is the indicator of tutor ${t}$ treatment status. We clustered standard errors at the student-tutor pair to account for repeated interactions:

\begin{align}
    \text{logit(Pr}(C_{mst}=1)) = \alpha_0 + \alpha_1 \text{Treatment}_{t} 
    \label{eq:strategy}
\end{align}

\paragraph{RQ4: Tutors' Perception of Tutor CoPilot.} To understand the treatment tutors' experiences and gather feedback on Tutor CoPilot, we conducted interviews with 18 treatment tutors a week after the study concluded.
These interviews explored their usage of Tutor CoPilot and how their use evolved over time.
We report the qualitative findings and themes that emerged from these interviews. 
Additional details on the interview setup are available in the Supplementary Information section~\ref{app:tutor_interview}.

\section*{Results}
We conducted a randomized controlled trial (n=783 tutors) in partnership with FEV Tutor, a virtual tutoring provider, and a large southern U.S. school district to evaluate whether Tutor CoPilot can improve tutoring quality and student outcomes. Tutors in the treatment group received access to Tutor CoPilot during their live tutoring sessions from March 27th to May 17th of 2024 (Methods).
Our theory of change centers on the observation that many tutors lack the experience to provide effective responses in the moment. 
As a result, they generate low-quality responses, such as giving generic feedback (e.g., ``Good job'') or giving away the answer after a single student attempt, which can hinder student learning. 
By offering real-time guidance tailored to the conversation and following high-quality strategies, Tutor CoPilot empowers tutors with more effective language to develop the student’s understanding. 
Thus, the tool builds tutors' skills while directly improving learning interactions. 

We report the results of our preferred model specifications for each research question below. 
Supplementary Information section~\ref{app:descriptive} reports descriptive results, such as the descriptive student numbers and tutor adoption of Tutor CoPilot; Supplementary Information section~\ref{app:student_level} reports the student-level analysis; Supplementary Information section~\ref{app:tutor_level} reports the tutor-level analysis; and Supplementary Information section~\ref{app:compliance} reports notes on compliance.

\subsection*{RQ1: To What Extent Does Tutor CoPilot Affect Student Learning?}

Table~\ref{tab:itt} shows the results of our intent-to-treat analysis, which includes estimates of the effect of the tutor having access to Tutor CoPilot on session-level outcomes using our preferred model specification. The model includes student-level covariates (baseline MAP math scores and demographic characteristics), school and grade fixed effects, and clusters residuals at the student-tutor pair level to account for correlated observations when students have multiple sessions with the same tutor.
 We provide Romano-Wolf adjusted p-values \citep{clarke2021romano} to control the inflated type I error rate from multiple hypotheses testing, and present alternative model specifications in Supplementary Information section~\ref{app:alternative_specs}.
``Participation points'' refers to points awarded by the tutor based on student engagement and effort during the session.
``Exit Ticket Attempted'' indicates whether the student attempted the exit ticket during the session. ``Exit Ticket Passed (Conditional)'' indicates whether the student passed the exit ticket, limited to sessions in which it was attempted. ``Exit Ticket Passed (Unconditional)'' captures whether the student passed without restricting to attempted sessions (i.e., students who did not attempt are counted as not passing).


\begin{table}[ht]
\centering
\resizebox{\textwidth}{!}{%
\def\arraystretch{1.15}
\begin{tabular}{lccccc}
\toprule
 & \multicolumn{5}{c}{\textbf{Panel A. Session outcomes}} \\
\cmidrule(lr){2-6}
& \textbf{Participation} & \textbf{Participation} & \textbf{Exit Tickets} & \textbf{Exit Tickets} & \textbf{Exit Tickets} \\
& \textbf{Points} & \textbf{Points Standardized} & \textbf{Attempted} & \textbf{Passed (Cond.)} & \textbf{Passed (Uncond.)} \\
\midrule
Treatment & 0.094 & 0.010 & 0.019$^{+}$ & 0.031$^{*}$ & 0.040$^{**}$ \\
& (0.27) & (0.028) & (0.011) & (0.014) & (0.015) \\
\hline
Control Mean & 14.071 & 0.016 & 0.843 & 0.732 & 0.617 \\
& (0.196) & (0.020) & (0.008) & (0.010) & (0.010) \\
Romano-Wolf p-val & [1.000] & [0.990] & [0.129] & [0.040] & [0.010] \\
N & 4136 & 4136 & 4136 & 3521 & 4136 \\
\midrule
 & \multicolumn{5}{c}{\textbf{Panel B. Student survey outcomes}} \\
\cmidrule(lr){2-6}
& \shortstack{My Tutor cared about \\ understanding math \\over memorizing the solution} & \shortstack{My tutor cared about \\how well I do in math} & \shortstack{Even when math is \\hard, I know I can learn it} & \shortstack{Session \\Rating} & \shortstack{Tutor \\Rating} \\
\midrule
Treatment & -0.0038 & 0.025 & 0.017 & -0.0015 & 0.026 \\
& (0.055) & (0.052) & (0.055) & (0.036) & (0.039) \\
\hline
Control Mean & 4.188 & 4.306 & 4.238 & 4.765 & 4.740 \\
& (0.038) & (0.035) & (0.037) & (0.025) & (0.027) \\
Romano-Wolf p-val & [1.000] & [0.950] & [1.000] & [1.000] & [0.812] \\
N & 1931 & 1931 & 1931 & 1948 & 1952 \\
\bottomrule
\end{tabular}}
\vspace{.5em}
\caption{Intent-to-treat analysis on student session-level outcomes. Estimates are from our primary model, which controls for baseline math scores, student demographics, and fixed effects for strata (school $\times$ grade). Participation points are standardized within-sample and by grade. Survey items are on a 5-point scale where higher is better. $^{+}p<0.1$; $^{*}p<0.05$; $^{**}p<0.01$; $^{***}p<0.001$ correspond to student-tutor pair clustered standard errors in parentheses. Romano-Wolf adjusted p-values are presented in brackets.  \label{tab:itt}}
\end{table}


We observed a significant positive treatment effect on students passing their exit tickets: 
Students working with treatment tutors were 4 percentage points (p.p.) more likely to pass their exit tickets ($p<0.01$, $62\% \rightarrow 66\%$ student passing rate from control to treatment). Supplementary Information section~\ref{app:tot} reports treatment-on-the-treated results using a two-stage least square instrumental variable model to estimate the impact of  \textit{using} Tutor CoPilot during the session, rather than just having access to it, showing students were 14 p.p. more likely to pass their exit tickets  ($p<0.01$) on average. These results remain significant under the Romano-Wolf multiple-hypothesis correction and under different model specifications presented in Supplementary Information section~\ref{app:alternative_specs}. 

We also found that conditional on attempting an exit ticket, students were 3 p.p. more likely to pass it ($p<0.05$) when they worked with a tutor with access to Tutor CoPilot. 
Our estimates also suggest students were 2 p.p. more likely to attempt the exit ticket during the session, but this result is not significant once we account for the multiple hypothesis correction. 

We also report treatment effects on participation points and student survey outcomes, neither of which are statistically significant. Although we pre-registered these outcomes and hypothesized positive impacts from Tutor CoPilot, our analysis did not find evidence supporting this expectation. Our study design anticipated that students would frequently match with the same tutor, but this occurred less often than planned. Frequent changes in tutor-student pairings might have made it difficult to influence the student perceptions measured by our surveys, compared to a model where students consistently interact with the same tutor across multiple sessions. 

\vspace{-1em}
\subsection*{RQ2: Does the Effect of Tutor CoPilot on Student Learning Differ by the Initial Effectiveness of Tutors?}

\begin{figure*}[ht]
    \centering
    \begin{subfigure}[b]{0.48\textwidth}
        \centering
        \includegraphics[width=\textwidth]{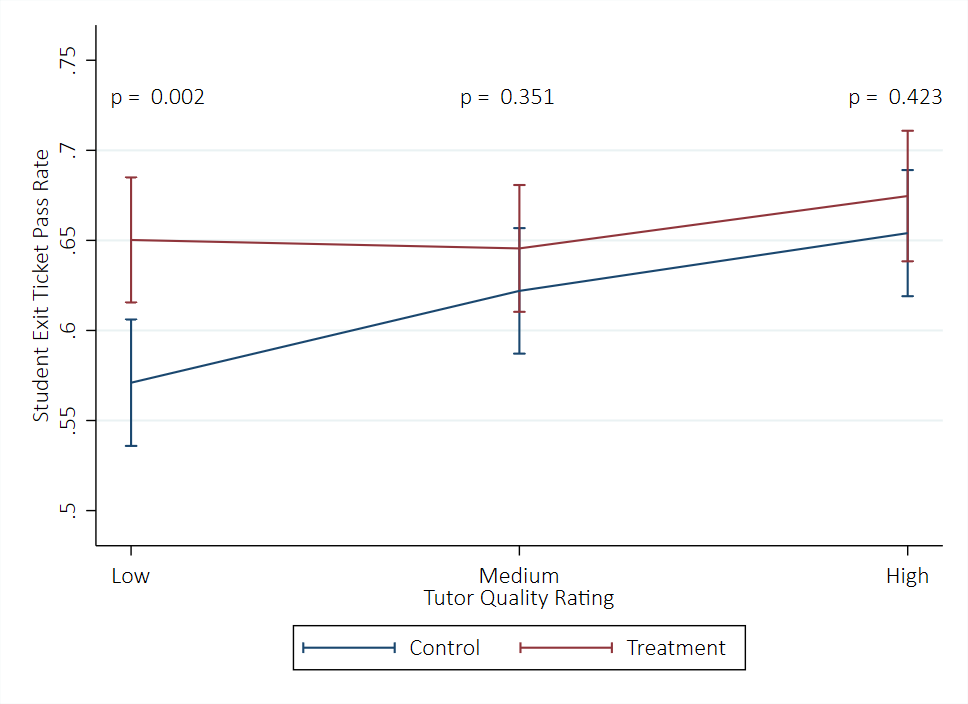}
        \caption{Tutor Quality Rating \label{fig:tutor_quality}} 
    \end{subfigure}
    \hfill
    \begin{subfigure}[b]{0.48\textwidth}  
        \centering 
        \includegraphics[width=\textwidth]{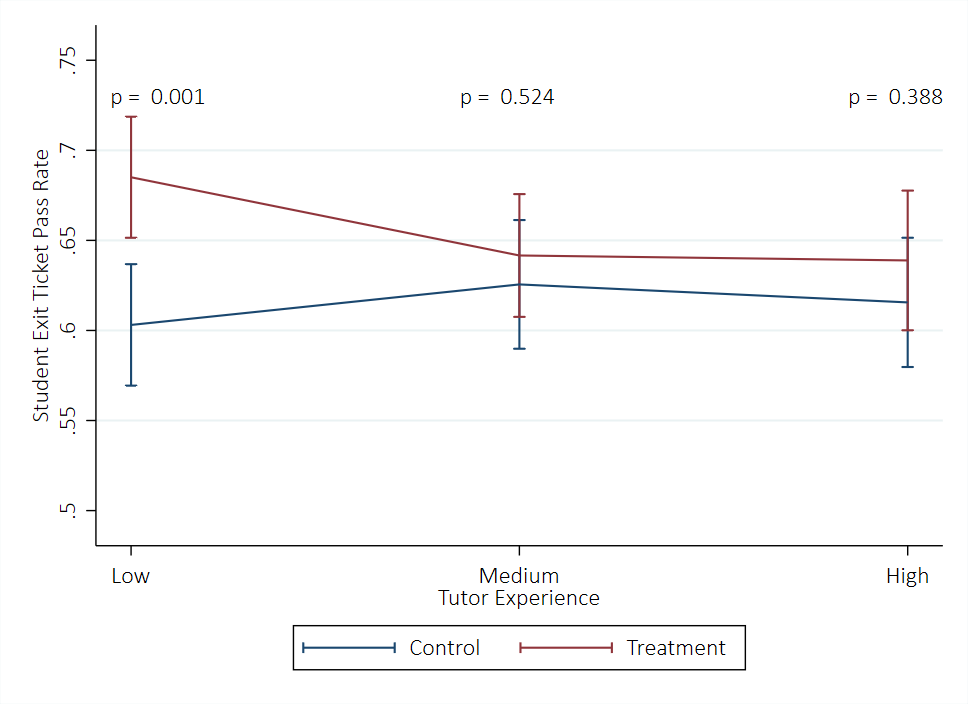}
        \caption{Tutor Experience \label{fig:tutor_experience}}  
    \end{subfigure}
    \caption{Heterogeneity analysis by tutor initial effectiveness on student learning. 
   Panel (a) reports by the difference by tutor's initial quality rating and panel (b) by tutor's tutoring experience. P-values reported for the difference between treatment and control tutors in the same category. \label{fig:tutor_effectiveness}}
\end{figure*}

Figure~\ref{fig:tutor_effectiveness} shows how the effect of Tutor CoPilot on student learning varies based on tutors' initial effectiveness, measured by their quality rating and level of experience at the tutoring platform.
The results indicate substantial benefits for tutors with lower initial effectiveness in both categories.
In Figure~\ref{fig:tutor_quality}, lower-rated tutors experienced a 9 p.p. increase in students passing their exit ticket ($56\% \rightarrow 65\%$ student passing rate from control to treatment).
In Figure~\ref{fig:tutor_experience}, less experienced tutors showed a 7 p.p. increase ($61\% \rightarrow 68\%$).
We saw these treatment effects diminish with increasing tutor effectiveness, but notably, students of lower-rated or less experienced tutors in the treatment group performed at or above the level of students with higher-rated or more experienced tutors in the control group. 
This suggests that Tutor CoPilot helped less-effective tutors achieve outcomes comparable to their more-effective peers.

\begin{table}[h!]
\centering
\resizebox{\textwidth}{!}{%
\def\arraystretch{1.15}
\begin{tabular}{ccccc}
\hline
\textbf{Quality}   & \textbf{Strategy Name}            & \textbf{Definition}           & \textbf{Examples}     & \textbf{F1} \\ 
\textbf{Category}   & \textbf{(Frequency)}              &            &    &  \\ \hline
\textcolor{cadmiumgreen}{\textbf{High}} & Prompt Student        & The tutor prompts the student     & ``Go ahead and try to explain & 0.89 \\
                            & to Explain                        & to explain a concept, rule, or    &  how you got the answer.'' & \\ 
                            &  (3\%)                            & their reasoning. \\ \hline
\textcolor{cadmiumgreen}{\textbf{High}} & Ask Question          & The tutor asks the student a      & ``What number can we multiply &  0.90  \\ 
                            & to Guide Thinking                 & question to help them think       &  the number 10 to get an equal \\
                            & (4\%)                                  & the problem.                      &  value of 100?'' \\ \hline
\textcolor{cadmiumgreen}{\textbf{High}} & Affirm Student's                  & The tutor affirms the student's   & ``Yes, 20 is the correct answer.'' & 0.65 \\ 
                            & Correct Attempt                 &  correct attempt. \\ 
                            & (13\%) \\\hline
\textcolor{cardinal}{\textbf{Low}}        & Ask Student to Retry              & The tutor asks the student to     & ``Please recheck your answer.''   & 0.73 \\
                            &    (1\% )                               & recheck their work or try again.  &  & \\ \hline
\textcolor{cardinal}{\textbf{Low}}        & Provide the Answer         & The tutor provides the final     & ``So, the greatest number will    & 0.76 \\ 
                            & or Explanation             &   answer or explanation to    &  be 7520.''   &   \\ 
                            &  (6\%) &  getting the final answer. \\ \hline
\textcolor{cardinal}{\textbf{Low}}        & Provide a Problem-           & The tutor provides a strategy or     & ``We can order the list according &  0.76 \\
                            & Specific Solution Strategy                & next step for solving the problem.          & to the hundredths place value.'' \\ 
                            &  (13\%) \\ \hline
\textcolor{cardinal}{\textbf{Low}}        & Encourage Student                 & The tutor encourages the          & ``That's a good try!'' & 0.81 \\
                            &  in Generic Way                   & student without being specific    &   &  \\
                            &  (12\%)                                & about the student's attempt.\\ \hline
\end{tabular}
}
\vspace{1em}
\caption{Taxonomy of \textcolor{cadmiumgreen}{high}- and \textcolor{cardinal}{low}-quality strategies, including their definitions, examples, and frequency over the labelled dataset of 241,066 messages sent by tutors during tutoring sessions. We train binary classifiers to identify these strategies at scale and report their test F1 score as well. A tutor response can include multiple strategies.    \label{tab:strategies}}
\end{table}

\vspace{-1em}
\subsection*{RQ3: How Does Tutor CoPilot Change Tutoring Quality as Measured by the Language Tutors Use with Students? }

We investigated whether treatment tutors employed higher-quality instructional strategies compared to control tutors by leveraging NLP methods, in a three-step process. 
First, we defined the taxonomy of strategies based on Wang et al. (2024a)\citep{wang-etal-2024-bridging} and salient response patterns in our data.  
Table~\ref{tab:strategies} reports the final taxonomy.
Second, we trained machine learning classifiers to identify these strategies at scale (Supplementary Information section~\ref{app:nlp_classifiers}). Finally, we used these classifiers to identify the strategies in our entire dataset of 350,000+ messages and estimated the difference in strategy use between the two groups using a logit regression. We clustered the standard errors at the student-tutor pair to account for the correlation between same-pair sessions. See Supplementary Information section~\ref{app:tutor_logodds} for odds ratios and Romano-Wolf p-values and alternative model specification with student and tutor random effects.

\begin{figure}[h!]
    \centering
    \includegraphics[width=0.9\linewidth]{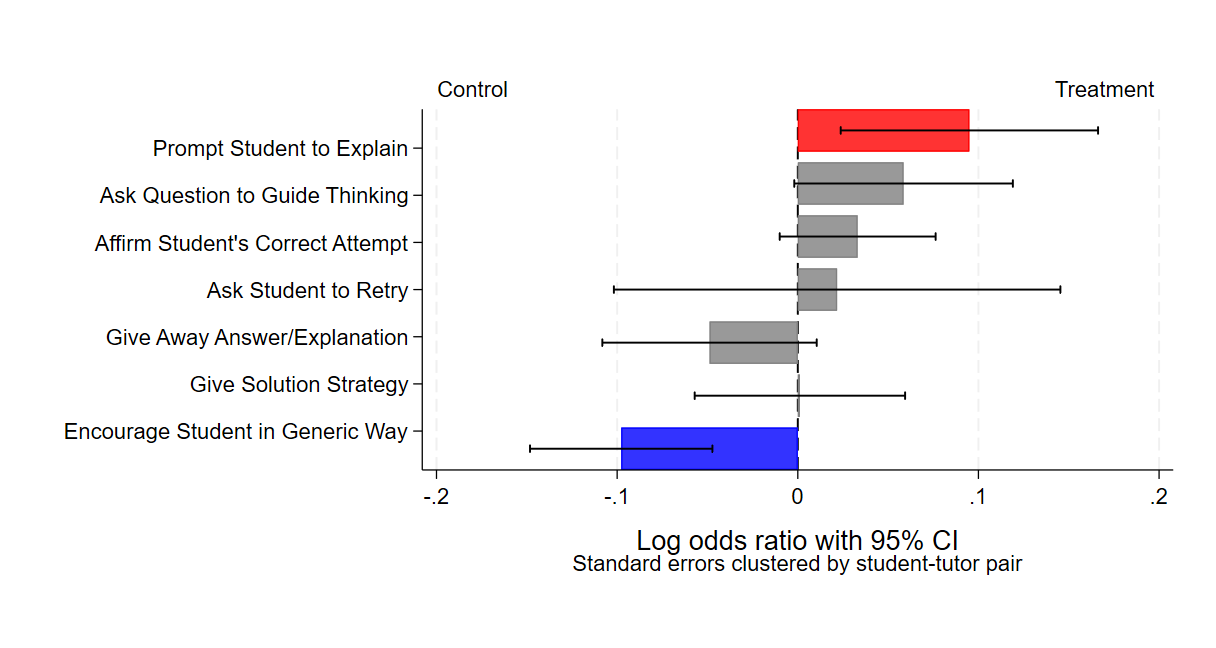}
    \caption{Strategies more likely to be used by control tutors (left) vs. treatment tutors (right). Logit regression coefficient results with 95\% Confidence intervals based on student-tutor pair standard errors and a sample of 241,005 classified tutor messages (115,517 from treatment and 125,488 from control tutors). Strategies more likely to be used by treatment tutors are shaded in red and strategies more likely to be used by control tutors are shaded in blue ($p<0.05$). Strategies not significantly different in use by either group are shaded in gray ($p>0.05$). See Table~\ref{tab:strategies_clustered} in Supplementary Information~\ref{app:tutor_logodds}.}
    \label{fig:strategies_clustered}
\end{figure}

Figure~\ref{fig:strategies_clustered} reports the log-odds of strategies used between treatment and control tutors, showing the impact of Tutor CoPilot on the types of strategies employed during tutoring sessions.

Based on our preferred specification, we found that treatment tutors were more likely to use strategies such as ``prompting the student to explain'', which is aligned with expert-recommended practices for promoting deeper learning \cite{chi1989selfexplanation,lepper2002wisdom}. This strategy was used approximately 10\% more in treatment sessions compared to control sessions. 
In contrast, control tutors were more likely to ``encourage students in a generic way'' compared to treatment tutors, using this strategy 10\% more. The higher likelihood of treatment tutors ``prompting the student to explain'' and control tutors ``encouraging students in a generic way'' are robust across all specifications and multiple-hypotheses testing adjustments.
These findings provide evidence that treatment tutors might have achieved better student learning outcomes through the use of more expert-like teaching strategies (see Supplementary Information section~\ref{app:tutor_logodds} for differences by tutor quality and experience).

\vspace{-1em}
\subsection*{RQ4: How Do Tutors Perceive Tutor CoPilot?} 
Interviews (n=18) indicated that tutors generally found Tutor CoPilot helpful, particularly for its ability to provide well-phrased explanations and break down complex concepts on the spot. 
They highlighted its usefulness in explaining difficult topics, such as differentiating between mean and median, and appreciated the clear definitions and hints it generated. 
However, some tutors noted areas for improvement, mentioning that the tool’s suggestions occasionally lacked alignment with students' grade levels. 
A common issue was that the responses were sometimes perceived as “too smart,” requiring tutors to simplify and adapt them for clarity. 
More details on the tutor interviews can be found in Supplementary Information section~\ref{app:tutor_interview}.

\section*{Discussion}

\subsection*{Background}
\paragraph{Training Novice Tutors to Use Effective Strategies.} One-on-one tutoring is one of the most effective ways to improve student outcomes and reduce educational disparities ~\citep{nickow2024impressive,dietrichson2017academic}, particularly in mathematics---a subject strongly linked to college graduation rates and future earnings~\citep{dougherty2003numeracy, watts2020academic,murnane2000important}. Expert tutors leverage deep pedagogical and content knowledge to adaptively respond to individual students needs~\citep{lepper2002wisdom, chi2001}. They see mistakes as critical sites for learning \citep{boaler2013ability}. They can diagnose and remediate misconceptions with strategies such as Socratic questioning, tailored hints, prompting students to articulate their reasoning and self-explain, and providing elaborated, responsive explanations, all of which stimulate critical thinking and deeper understanding~\citep{ chi1989selfexplanation, lepper2002wisdom,webb1989, graesser1994, roscoe2007, lin2022goodmove,easley1975teaching}. Good math teaching can be especially challenging, as it requires instructors to engage students in reasoning about abstract concepts and solving complex problems~\citep{wood1976role, hattie2008visible}, as well as understanding the developmental progression of problem-solving strategies to interpret students' thinking on the fly~\citep{carpenter2003thinking}.

Training novice tutors to achieve this level of expertise is challenging. Traditional professional development programs (e.g. workshops) are often costly~\citep{heinrich2014improving}, time-intensive~\citep{kelly2020using}, and disconnected from the real-time dynamics of tutoring environments \citep{hill2020professional}, frequently limiting their effectiveness~\citep{KraftBlazarHogan2018, knight2012assessing}. As a result, students often work with novice tutors who struggle to remediate their mistakes in real time, and provide them with direct answers or overly generic feedback, missing out on critical learning opportunities~\citep{wang-etal-2024-bridging, lepper2002wisdom, boaler2013ability, graesser1994, roscoe2007}. For example, tutors at FEV---our study's context---would often respond to students with generic and scripted messages like ``Incorrect'', ``Please try again.'', ``The answer is ..." \cite{wang-etal-2024-bridging}. Our work seeks to address these challenges by providing a scalable, cost-effective solution that provides guidance to novice tutors in their live teaching contexts, helping them navigate the complexities of teaching mathematics. This approach also offers infrastructure for conducting experimental studies on effective tutoring practices.

\paragraph{AI in K-12 Education.} Recent advances in AI have sparked excitement about the potential for LLMs to transform K-12 education, from intelligent tutoring systems (ITS) to automated feedback generation~\citep{ khan_academy_2023, graesser2004autotutor}. LLMs have the potential to produce adaptive, human-like responses, enhancing earlier generations of ITSs. However, tuning LLMs to behave like expert teachers is challenging. Out-of-the-box LLMs are trained predominantly on internet text (e.g., Wikipedia) that differs greatly from authentic teacher-student interactions. Even strong prompting or fine-tuning techniques struggle to bridge this gap ~\citep{wang2023a, Singer_2023, ji2023survey, frieder2023mathematical}, as they tend to rely on surface-level patterns rather than deeper pedagogical reasoning ~\citep{jurenka2024towards, beurer2023prompting, mccoy2023embers} that expert educators develop through experience~\citep{polanyi1966logic,  seamster1993cognitive}. 
As a result, LLMs often fail to use pedagogical strategies that foster deeper understanding~\citep{wang2023a, Singer_2023, frieder2023mathematical}, like probing questions and others listed above, which can hinder students' educational outcomes~\citep{bastani2024generative, nie2024gpt}. Recent efforts have identified approaches to improving LLMs ability to produce expert-like responses in education, e.g., by providing editable templates \cite{lin2024can} and pedagogical instruction following \cite{team2024learnlm}.
We build on prior work with Bridge~\citep{wang-etal-2024-bridging}, an adaptation method that captures expert decision-making by transforming raw think-aloud data with experienced educators into effective instructions for LLMs. 
Bridge has been validated to outperform fine-tuning and prompt-engineering baselines in generating expert-like pedagogical responses.

\vspace{-1em}
\paragraph{Approaches to Human-AI Collaboration.} Human-computer collaborative systems combine the complementary strengths of humans and computers to enhance performance across various domains like healthcare~\citep{lai2021human}, writing~\citep{ lee2022coauthor} and data annotation~\citep{kim2024meganno+}. In education, this collaboration means using AI to support human expertise \emph{in situ} rather than replacing it.
Prior research shows AI assisting in discrete tasks like patient treatment selection ~\cite{cai2019hello,mosquera2014computer}, and guiding interactions in more structured, scripted environments such as call centers, where it provides novice workers with immediate, context-specific suggestions ~\citep{brynjolfsson2023generative}. However, far less work has addressed AI assistance for complex interaction tasks like teaching or tutoring, which entail dynamic, open-ended dialogues requiring AI to work in tandem with a human tutor over extended exchanges. Educators bring grounded, context-specific knowledge and empathy that AI lacks, while AI can speedily access vast information to offer real-time guidance. This complementary setup can preserve the critical human connection in teaching \cite{sabol2012relationships,robinson2022framework} mitigate LLM lapses through human oversight---and is thus recommended by many in the field of AI and education \cite{nguyen2023ethical,holstein2019designing,brusilovsky2024ai}. 
Drawing on prior work~\citep{wang-etal-2024-bridging}, we emphasize the human's crucial role in selecting expert-curated strategies to guide LLMs in generating effective pedagogical suggestions. 
Prior findings suggest that without human selection, LLMs often default to repetitive strategies, whereas experienced educators employ richer, more diverse teaching strategies~\citep{wang-etal-2024-bridging}. 
This observation underscores the human's critical role in guiding AI to deliver contextually appropriate and impactful instructional support.

\subsection*{Implications of Tutor CoPilot Findings}
This study introduces Tutor CoPilot, a human-AI tool to scale real-time expertise in K-12 tutoring. 
It represents the first randomized controlled trial of AI-supported live tutoring. 
Our findings demonstrate that AI-generated guidance---based on expert thinking---can significantly improve tutoring quality, particularly for less experienced tutors. 
Notably, Tutor CoPilot improves student learning at a low cost of \$20 per tutor annually, which is far cheaper than traditional training programs costing thousands of dollars.
We also find that Tutor CoPilot helps tutors adopt high-quality strategies that promote deeper learning. Thus, our approach offers a practical and scalable solution for improving education, especially for under-served communities that stand to benefit the most from better quality education.  

Tutor CoPilot also provides infrastructure for gathering experimental evidence on effective tutoring practices. It allows researchers to answer the question: how does suggesting a particular set of practices to tutors affect student learning? Our results show that students master lessons faster when their tutors are offered a bundle of expert-preferred strategies for mistake remediation. Evidence from this experimental method can support the design of effective tutor training programs.

\subsection*{Limitations} This study presents promising results on Human-AI approaches for scaling real-time expertise. However, we note the following key limitations to our work: generalizability, effectiveness of strategies, dosage and exposure, multi-modality, and user safety.

\vspace{-1em}
\paragraph{Generalizability.} First, our intervention on Tutor CoPilot is conducted with novice tutors who teach young students from under-served communities (majority Hispanic) based in a Southern school district in the United States. Tutor CoPilot may be less effective for experienced educators, for whom generated suggestions may offer limited benefit \cite{wang-etal-2024-bridging}.
The learning needs of these students may differ significantly from those of students with different demographics, in other parts of the United States, in other countries, or in more affluent learning environments. 
Additionally, while we observed substantial improvements in students' proximal learning gains (i.e., the students' exit ticket performance), we did not find statistically significant improvements in end-of-year math test scores; please refer to Supplementary Information section~\ref{app:student_level} for this analysis.
This may be due to the little variation in students' exposure to the treatment and the relatively short duration of the study (two months). 
Future research would benefit from randomizing students to better assess the impact of the intervention. Although our human-AI collaborative approach has broader potential across domains that rely on human expertise, such as law and healthcare, evaluating this potential requires careful adaptation and validation to meet the demand of those professions.

\vspace{-1em}
\paragraph{Effectiveness of Strategies.} Our study provides suggestive evidence of the impact of pedagogical strategies on student learning. Tutor CoPilot offers a bundle of strategies from which tutors can choose and edit the suggested responses. We are measuring the causal impact of offering the tool, rather than that of using particular strategies. We opted for this design, in which tutors have the flexibility to choose, to utilize tutor knowledge of the context (in some contexts some strategies may be more effective than they are in other contexts) and to maintain tutor agency, which alone can be important for their learning \citep{smith2017teachers, brod2023agency, calvert2016power, schon2017reflective, vahasantanen2017professional}. Future work might create experimental designs that allow for testing the impact of strategies individually while maintaining tutor agency and for testing the effects of the tool on tutor learning and effectiveness. Such work could also consider measuring key contextual factors that may influence the effectiveness of a strategy at a given moment (e.g., the number and types of attempts a student has made at solving a problem).

\vspace{-1em}
\paragraph{Dosage and Exposure.} Our study design had to account for the tutoring platform's almost-random algorithm to match tutors to students each session, which led to our choice of randomizing tutors and focusing on session-level outcomes. Combined with a short implementation period of about 2 months, most students in our sample had very little exposure to tutors with access to the tool that alternated with control tutors, which may also have restricted the effect of the tool on student perceptions measured by our survey and End-of-Year MAP achievement scores. Future research will benefit from a design that allows for higher intensity and contrast of Tutor CoPilot's exposure between students to measure the tool's impact on achievement scores and consistent student-tutor matching to measure changes in student perceptions of their tutors. 

\vspace{-1em}
\paragraph{Modality.} Our study collaborates with a chat-based tutoring platform, which is well-suited for real-time use of language models. 
Future work would benefit from extending Human-AI approaches to include other modalities---such as vision (e.g., whiteboard or student face) and speech (e.g., student and tutor voice)---to incorporate naturalistic modalities for interaction.
Including these modalities would provide a more comprehensive understanding of the student and thus enable better expert-like guidance during the interaction.
The shift towards multi-modal approaches also raises interesting challenges on how to provide real-time guidance through different modalities, such as via whiteboard drawing or verbal explanations. 

\vspace{-1em}
\paragraph{User safety.} While our study de-identifies names, user identities can still be inadvertently disclosed through non-name information, such as email addresses, phone numbers, or personal anecdotes shared by students and tutors. 
Similar privacy concerns arise when incorporating vision and speech modalities, highlighting the need for better safeguards.
Additionally, we observe a trade-off between performance and safety. 
In our study, to mitigate over-exposure, we limited the conversation context passed to external APIs; in initial pilot studies, we varied the conversation context between 5 to 10 messages and found that tutors preferred the quality of interactions when more context was included. 
Balancing these factors will be important for future research.

\subsection*{Future Work}
Looking ahead, our work can extend in several exciting directions.
First, we aim to explore how well novices retain the skills they acquired from real-time expert guidance. 
This analysis will shed light on the long-term effects of the use of AI systems in real-world settings.
Second, expanding Tutor CoPilot to other skill areas---such as developing collaborative learning---and applying it in different subjects, age groups, or system designs will provide insights into its broader impact. 
Last, we observed significant variation in how tutors adapted or personalized AI-generated suggestions,  highlighting interesting human-AI collaborative dynamics that merit further study.

\section*{Data Availability}
Student data are protected by a data sharing agreement with the participating district. The datasets analyzed for this study are available from the corresponding author upon request.

\section*{Code Availability}

Code to produce the analysis and resulting figures and tables presented in this study is available at \url{https://github.com/anatrindaderibeiro/tutorcopilot_analysis}. The NLP classifiers developed for our analysis of the use of strategies by tutors and to understand the tutoring moments when Tutor CoPilot was activated can be found at \url{https://huggingface.co/StanfordSCALE}.
We also provide a tutorial of the Tutor CoPilot tool at \url{https://github.com/rosewang2008/tutor-copilot/tree/main}.


\end{bibunit}

\section*{Acknowledgments}
We thank Jonathan Bechtel, Martin Viau, Daniel Hebert, Shafiq Ahmed, Vivek Patel for their guidance and support in this study; Allen Nie, Greg Stoddard, Xuechen Li, Emma Brunskill, Kristina Gligorić, Megha Srivastava, Aishwarya Mandyam, and Chenglei Si for insightful discussions and pointers; participants at UChicago's Becker-Friedman Institute AI for Social Science conference, the American Economic Association Annual Meeting, and The Society for Research on Educational Effectiveness Conference for their helpful feedback; and Xander Beberman for excellent support during the revision process.

\section*{Author Contributions}
R.E.W. and A.T.R. led the study design, data analyses, and writing of the manuscript; R.E.W. conceived the tool implementation; C.R., S.L., and D.D. were involved in every phase of the study, particularly the conception of the study, the study design, the preparation of study materials, the interpretation of analyses, and the writing of the manuscript.

\section*{Competing Interests}
Authors A.T.R., C.R., S.L. and D.D. declare no competing interests. R.E.W. worked part-time with the tutoring provider to build Tutor CoPilot into the tutoring platform.

\section*{Tables}


\begin{table}[h!]
\centering
\resizebox{\textwidth}{!}{
\def\arraystretch{1.15}
\begin{tabular}{l*{7}{c}}
\toprule
 & \shortstack{ Prompt \\ Student \\ to Explain} & \shortstack{Ask \\ Question \\ to Guide \\ Thinking} & \shortstack{Affirm \\ Student's \\ Correct \\ Attempt} & \shortstack{Ask \\ Student \\ to Retry} & \shortstack{Give \\ Away \\ Answer/ \\ Explanation} & \shortstack{ Give \\ Solution \\ Strategy} & \shortstack{Encourage \\ Student in \\ Generic \\ Way}    \\
\midrule
Treatment           &       0.095\sym{**} &       0.059\sym{+}  &       0.033         &       0.022         &      -0.049         &      0.0012         &      -0.098\sym{***}\\
                    &     (0.036)         &     (0.031)         &     (0.022)         &     (0.063)         &     (0.030)         &     (0.030)         &     (0.026)         \\
\hline
Romano-Wolf p-val   &     [0.010]         &     [0.020]         &     [0.158]         &     [0.871]         &     [0.129]         &     [0.990]         &     [0.010]         \\
Z                   &       2.614         &       1.897         &       1.505         &       0.345         &      -1.614         &       0.039         &      -3.795         \\
Odds Ratio          &       1.100         &       1.060         &       1.034         &       1.022         &       0.952         &       1.001         &       0.907         \\
N                   &      241005         &      241005         &      241005         &      241005         &      241005         &      241005         &      241005         \\

\bottomrule
\end{tabular}
}
\vspace{1em}
\caption{  Log Odds Analysis of Strategy Use Differences Between Treatment and Control Tutors. The parentheses report the standard error
clustered by student-tutor pairs. Romano-Wolf adjusted p-values are shown in brackets.  $^{+}p<0.1$; $^{*}p<0.05$; $^{**}p<0.01$; $^{***}p<0.001$. \label{tab:strategies_clustered}}
\end{table}



\clearpage
\appendix


\onecolumn

\begin{bibunit}

\newcommand{\minilogo}{
    \includegraphics[scale=0.15]{figures/logo.png}}
    
\newcommand{\pencil}{
    \includegraphics[scale=0.09]{figures/emojis/pencil.png}}

\newcommand{\refresh}{
    \includegraphics[scale=0.09]{figures/emojis/refresh.png}}
    
\newcommand{\dropdown}{
    \includegraphics[scale=0.09]{figures/emojis/dropdown.png}}

\begin{center}

\section*{Supplementary Information}

\section*{Tutor CoPilot: A Human-AI Approach for  Scaling Real-Time Expertise}
\centering Rose E. Wang, Ana T. Ribeiro, Carly D. Robinson, Susanna Loeb, Dora Demszky

\vspace{1in}
\end{center}

\section{Tutor CoPilot \label{app:tutor_copilot}}
Here we describe how Tutor CoPilot works on the  \textit{front end} (what the tutor sees) and the \textit{back end} (what our system does under the hood).\footnote{One of the authors of this study worked part-time with FEV Tutor, the tutoring provider, and collaborated closely with several teams (e.g., engineering, design, tutoring operations and curriculum teams) to build Tutor CoPilot and framework for LM-based interventions.
Supplementary Information section~\ref{app:pilot_studies} provides information on how we tested the Tutor CoPilot system prior to deployment.} 
Figure~\ref{fig:tutor_copilot} shows a high-level overview of Tutor CoPilot. 

Tutor CoPilot, as shown in Figure 1a, is directly integrated into tutoring sessions to support live interactions. 
The current setup embeds Tutor CoPilot within a virtual tutoring platform that features a problem display, shared whiteboard, and chat window. 
A new Tutor CoPilot button\minilogo{} is added, allowing tutors to easily activate it during sessions for immediate assistance that seamlessly fits into their workflow. 

Once activated, Tutor CoPilot, as shown in Figure 1b, pulls relevant information from the ongoing interaction, such as the conversation context, lesson topic and the requested strategy.\footnote{The strategy is chosen either from default options or a dropdown menu (Figure~\ref{fig:tutor_copilot}d).}
Currently, the conversation context is based on chat interactions, but the system is flexible and could ultimately be adapted to process speech or visual inputs, like whiteboard activity. 
To ensure user safety and privacy, we automatically de-identify student and tutor names retrieved from the roster database using placeholders ``[STUDENT]'' and ``[TUTOR]'' via Edu-ConvoKit \citep{wang2024convokit}. 
Additionally, when using external LM services (e.g., OpenAI), we limit shared conversation context to the 10 most recent messages to minimize data exposure. 
These measures lay the groundwork for future privacy safeguards with human-AI systems.

Leveraging Bridge~\citep{wang-etal-2024-bridging}, we generate expert-like suggestions, as shown in Figure 1c, based on the de-identified conversation, lesson topic, and the chosen strategy (e.g., ``provide a hint'') using OpenAI's GPT-4 model.
This approach \textit{lightens the cognitive load} on tutors by eliminating the need for them to figure out how to prompt the model themselves and allows them to focus on delivering high-quality instruction. Wang et al.~(2024a)\citep{wang-etal-2024-bridging} validates Bridge for this study's context (FEV tutor), showing that U.S.-based teachers rated responses generated by our deployed model higher than the original tutor responses across all quality dimensions (preference, usefulness, care, human-soundingness). Bridge, including the dataset and model prompts, is open-sourced for others to use. 

Tutors can personalize the generated guidance, as shown in Figure 1d, by editing the suggested responses (\pencil), regenerating them (\refresh), or selecting a different strategy (\dropdown). 
Available strategies include providing a solution, a worked example, a minor correction, a similar problem, simplifying the question, affirming the correct answer, and encouraging the student. 
Selecting a new strategy updates the suggestion in the response box.
Unlike typical autocompletion systems that provide a single response, Tutor CoPilot presents multiple suggestions of responses based on different strategies. 
By giving tutors a range of options to choose from, our approach helps maintain response quality by drawing on tutors' contextual knowledge and preserves tutor agency, which both increases effectiveness and supports professional development. \cite{smith2017teachers,brod2023agency,calvert2016power,schon2017reflective}.
\vspace{-1em}

\subsection{Pilot Studies of Tutor CoPilot \label{app:pilot_studies}}

Before we fully launched Tutor CoPilot, we conducted two pilot studies to ensure its usability and performed data pulls to ensure our databases accurately tracked information needed for our analysis. 
These pilot studies involved 10-20 tutors who were not part of the main intervention.
The pilot studies played a critical role in the success of Tutor CoPilot because they identified system bottlenecks that made Tutor CoPilot difficult to use. 
One main issue from the first pilot study was the response time of Tutor CoPilot: Data analyses, interviews and surveys with tutors surfaced frustrations that the tool took longer than 30 seconds to respond. 
We profiled Tutor CoPilot and identified inefficiencies in our data retrieval pipeline.
The second pilot study showed significant response time improvements and strong tutor satisfaction.
This gave us the green light to deploying the tool at scale for our study.

\section{Exit Ticket Significance and Expected Impact on MAP\label{app:exit_ticket}}


\begin{table}[h!]
\centering
\begin{tabular}{lc}
\toprule
\textbf{Dependent}      &   EOY MAP Math Std (2022-23sy)  \\
\midrule
Exit Tickets Passed    & 0.02*** \\
                            & (0.002) \\
BOY MAP Math Std (2022-23sy)         & 0.87*** \\
                            & (0.02) \\
\hline
R$^2$        & 0.728       \\

N                 &   689 \\
\bottomrule
\end{tabular}
\vspace{1em}
\caption{ Linear Regression End-of-Year Standardized MAP Math Scores on Exit Tickets Passed and Beginning-of-Year Standardized MAP Math Scores. This regression uses exit ticket and achievement data from students in the same district of our study who received tutoring services from FEV Tutor during the 2022-23 school year, before Tutor Copilot was implemented. Results represent the predictive value of student beginning of the year (BOY) MAP performance and the number of exit tickets successfully completed during the year for the end of the year (EOY) MAP performance. We used standardized BOY and EOY MAP performance measures and robust standard errors. $^{+}p<0.1$; $^{*}p<0.05$; $^{**}p<0.01$; $^{***}p<0.001$.
\label{tab:exit_ticket_control}}
\end{table}

We demonstrate the predictive value of exit tickets to students end of the year MAP Math achievement scores in Table~\ref{tab:exit_ticket_control}. We use complete session-level data with student exit ticket information for students in the district receiving FEV Tutor tutoring during the 2022-23 school year, when Tutor CoPilot was not available to tutors. We find that the exit ticket passing rate predicts students' end-of-year test scores (0.02, $p<0.001$), suggesting that improvements in short-term learning outcomes measured by the exit ticket are associated with improvements in long-term academic performance. 

Although our study design does not allow us to directly estimate the impact of Tutor CoPilot on MAP achievement scores, we can perform a back-of-the-envelope calculation of the expected impact of a full year of tutoring sessions with Tutor CoPilot enabled on student achievement scores using the predictive value of exit tickets on MAP scores and our estimated impact of Tutor CoPilot on exit ticket completions at the session level. Considering FEV Tutor's recommended dosage is three sessions per week for at least 10 weeks during the school year and that having Tutor CoPilot enabled for every session would increase the likelihood of students passing exit tickets by four percentage points per session, we should expect students to complete an additional ($0.04x30=$) 1.2 exit tickets per school year on average. Thus, a full year of tutoring with Tutor CoPilot enabled for tutors is expected to increase student performance on EOY MAP Math by 1.2 times the predicted value of exit tickets, 0.02 standard deviations, which is a 0.024 standard deviations gain on average. 

Furthermore, our Treatment-on-the-treated estimate, i.e., the effect of Tutor CoPilot for sessions in which it was used, suggests the upper bound effect of Tutor CoPilot access across all tutors to be an increase of up to ($0.14x30=$) 4.2 exit tickets during the school year, with an associated increase on EOY MAP Math of ($4.2x0.02sd=$) 0.084 standard deviations. For students of tutors who would be rated as less effective, this effect would be bigger - more than two times as large.

Based on the explanatory power of BOY MAP Math and other typical student-level covariates, a sufficiently powered experimental design to confirm an effect size of 0.024 standard deviations on EOY MAP Math would require approximately 16,000 students to be assigned to the treatment (tutoring with Tutor CoPilot enabled) or control (tutoring without Tutor CoPilot enabled) for a full school year of tutoring.

\section{Descriptive Statistics \label{app:descriptive}}

\paragraph{Study in numbers.}
Table~\ref{tab:session_statistics} reports the session-level statistics. A majority of sessions are with elementary school students, particularly those in Grade 4.
Table~\ref{tab:participant_statistics} reports student- and tutor-level statistics.

\begin{table}[h!]
\centering
\resizebox{0.45\textwidth}{!}{%
\def\arraystretch{1.15}
\begin{tabular}{ccccc}
\toprule
\multicolumn{5}{l}{\textit{Panel A. Elementary Schools}} \\
\midrule
            & Grade 3   & Grade 4   & Grade 5   & Total     \\
Total       & 676       & 1,828      & 357       &  2,861    \\
\midrule
\multicolumn{5}{l}{\textit{Panel B. Middle Schools}} \\
            & Grade 6   & Grade 7   & Grade 8   & Total     \\
Total       & 1,275       & 0      & 0       &  1,275  \\
\midrule 
\textbf{Total}       &      &       &        &  \textbf{4,136}  \\\bottomrule
\end{tabular}
}
\vspace{1em}
\caption{Session-level Statistics by Grade. We had 8 elementary schools and 1 middle school in our study sample. \label{tab:session_statistics}}
\end{table}

\begin{table}[h!]
\centering
\resizebox{0.65\textwidth}{!}{%
\def\arraystretch{1.15}
\begin{tabular}{lll}
\toprule
\multicolumn{2}{c}{}      & \textbf{Mean}  \\ \midrule
\textit{Panel A: Student Characteristics ($N = 1,013$)}    \\
Gender                              \\ 
                            & Male      & 0.46      \\
                            & Female    & 0.46      \\
                            & Missing   & 0.09      \\
Race/Ethnicity              \\
                            & Hispanic              & 0.791         \\
                            & White                 & 0.082         \\
                            & Black                 & 0.029         \\
                            & Asian                 & 0.003         \\
                            & Pacific Islander      & 0.001         \\
                            & American Indian or Alaska Native       & 0.001     \\
                            & Two or more races     &  0.008         \\
                            & Missing               &  0.086        \\
Free/Reduced Lunch  &           &           \\
                            &  No       &  0.89     \\
                            &  Yes      &  0.02     \\
                            &  Missing      &  0.09     \\
Limited English Proficiency Program       &           &       \\
                            & No        & 0.56     \\
                            & Yes       & 0.35     \\
                            & Missing   & 0.09     \\
Special Education Program       &           &       \\
                            & No        & 0.80     \\
                            & Yes       & 0.11     \\
                            & Missing   & 0.09     \\
                            & & \\
MOY Math MAP score  &       Mean    & 249 \\
                            & sd   & 16.9     \\
                            & Min   & 132     \\
                            & Max   & 249     \\
                            & Missing   & 0.15     \\

\hline
\textit{Panel B: Tutor Characteristics ($N = 783$)}    \\
Gender                              \\ 
                            & Male      & 0.45      \\
                            & Female    & 0.55      \\
                         & & \\
Quality Rating              &       Mean    & 0.41 \\
                            & sd   & 0.20     \\
                            & Min   & -0.33     \\
                            & Max   & 1.11     \\
                            & Missing   & 0.001     \\
                         & & \\
Experience                  &       Mean    & 21 \\
                            & sd   & 12.5     \\
                            & Min   & 8     \\
                            & Max   & 73     \\
                            & Missing   & 0     \\
\bottomrule
\end{tabular}
}
\vspace{1em}
\caption{Descriptive statistics for students and tutors who participated in the study. We define a participating student as a student who has at least one tutoring session recorded with a treatment or control tutor during the implementation period. We define a participating tutor as a tutor randomized to the treatment or control condition who has at least one tutoring session recorded during the implementation period with a student from the school district of interest.\label{tab:participant_statistics}}
\end{table}

\begin{figure}[h!]
    \centering
    \includegraphics[width=0.8\linewidth]{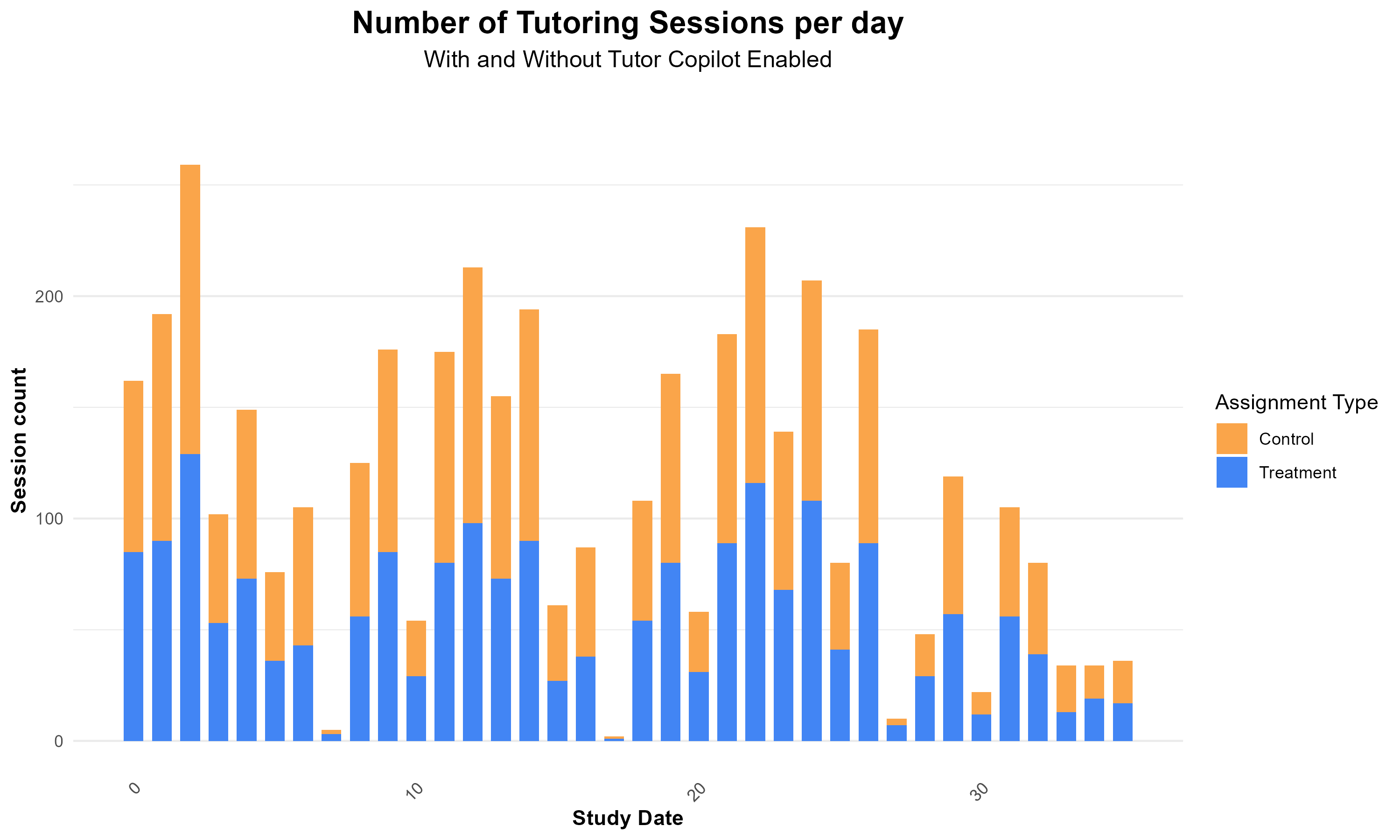}
    \caption{Number of sessions per day by tutor assignment. }
    \label{fig:tutor_copilot_sessions}
\end{figure}

\begin{figure}[h!]
    \centering
    \includegraphics[width=0.8\linewidth]{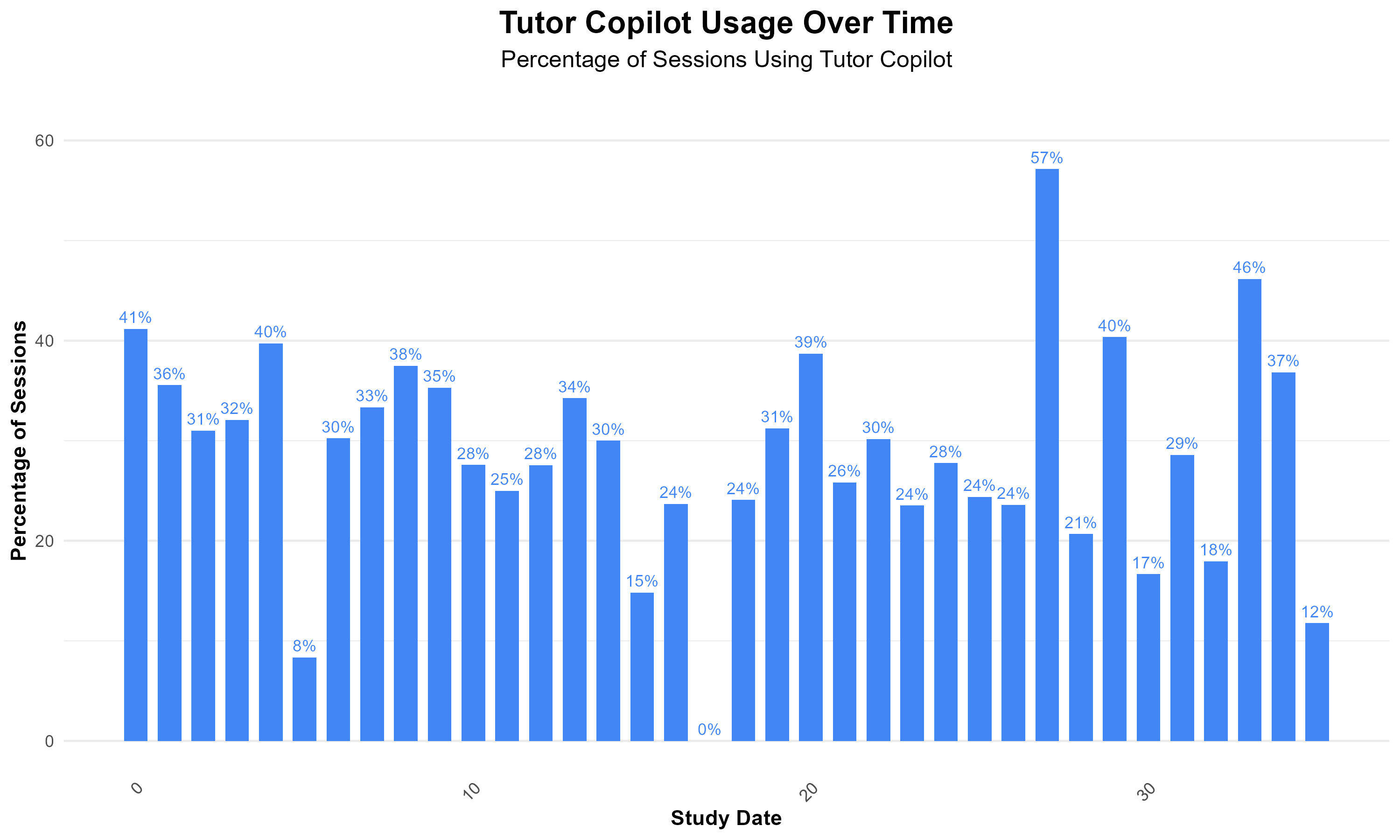}
    \caption{Percentage of treatment sessions that used Tutor CoPilot at least once in their session. About 29\% of treatment sessions used Tutor CoPilot during our study.}
    \label{fig:tutor_copilot_usage}
\end{figure}

\begin{figure*}[h!]
    \centering
    \begin{subfigure}[b]{0.80\textwidth}
        \centering
        \includegraphics[width=\textwidth]{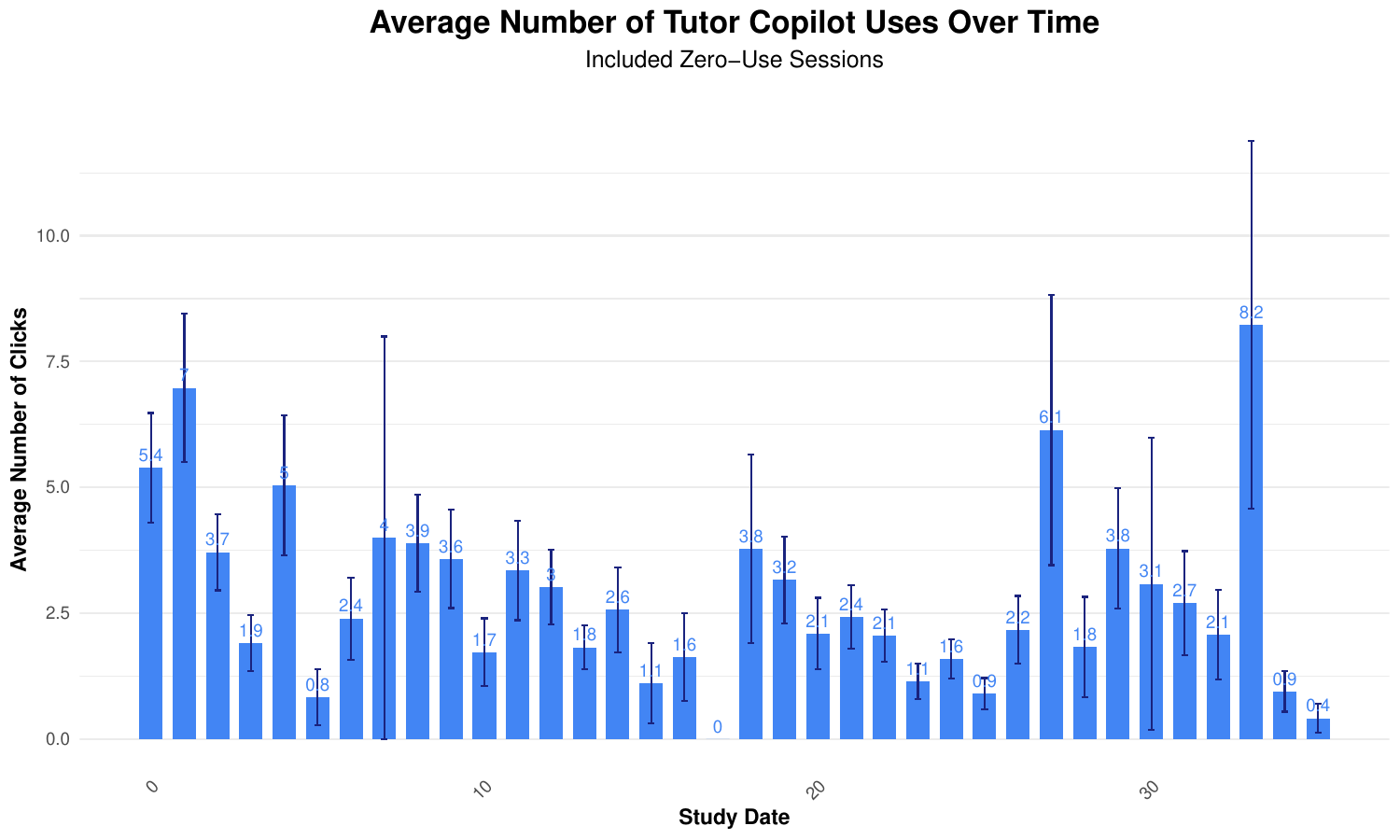}
        \caption{} 
    \end{subfigure}
    \hfill
    \begin{subfigure}[b]{0.80\textwidth}  
        \centering 
        \includegraphics[width=\textwidth]{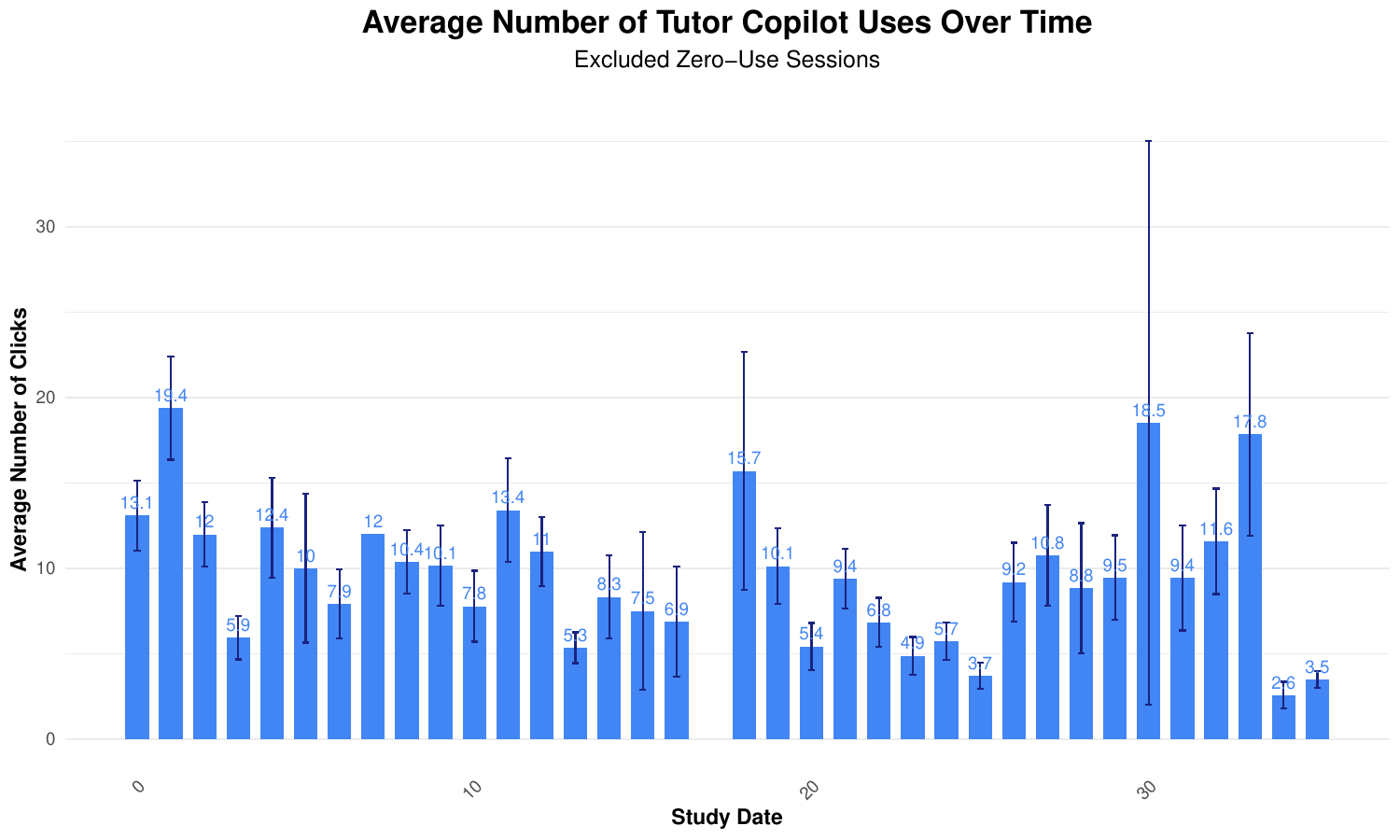}
        \caption{}  
    \end{subfigure}
    \caption{(a) reports the average number of Tutor CoPilot uses, including the sessions that had no usage, and (b) reports the average number of uses but excludes the sessions with no usage. When including the zero-use sessions, tutors use Tutor CoPilot about 3 times per session. When excluding the zero-use sessions, tutors use Tutor CoPilot about 10 times per session. \label{fig:tutor_copilot_clicks}}
\end{figure*}

\paragraph{Tutor CoPilot usage.} Figure~\ref{fig:tutor_copilot_sessions} reports the number of control and treatment tutoring sessions conducted per day during the implementation period. Figure~\ref{fig:tutor_copilot_usage} reports the percentage of treatment sessions that used Tutor CoPilot at least once in their session. On average, we find that about 29\% of treatment sessions used Tutor CoPilot. 
Figure~\ref{fig:tutor_copilot_clicks} reports the average number of uses per session, either (a) including or (b) excluding the sessions with zero uses. The tool was activated by 239 different treatment tutors in 599 of the 2,014 sessions with treatment tutors in our study. When including the zero-use sessions, treatment tutors used Tutor CoPilot about 3 times per session. When excluding the zero-use sessions, treatment tutors used Tutor CoPilot about 10 times per session. We did not send any reminders to tutors to use Tutor CoPilot. We also found that four different control tutors activated Tutor CoPilot in six of the 2,122 sessions with control tutors, with no significant impact to our estimates (see Supplementary Information about compliance in section \ref{app:compliance}) 

\paragraph{Moments of Tutor CoPilot usage.} We were interested in the moments of tutoring that tutors used Tutor CoPilot, e.g., did tutors use Tutor CoPilot at the start of the tutoring session or as the student started to solve problems? 
To study this, we developed a taxonomy of tutoring moments and trained binary classifiers to identify these tutoring moments. 
Supplementary Information section~\ref{app:nlp_classifiers} details development process in more depth. 
Table~\ref{tab:moments} describes the taxonomy of tutoring moments, the frequency of these moments in our final data, and the test F1 score of the corresponding moment classifier. 
We find that tutors use Tutor CoPilot mostly during the moments of student learning: when the student is attempting the problem, or after they have attempted the problem and the tutor is giving them feedback.


\begin{table}[h!]
\centering
\resizebox{\textwidth}{!}{%
\def\arraystretch{1.15}
\begin{tabular}{llll}
\hline
\multicolumn{1}{c}{\textbf{Moment (Frequency)}}            & \multicolumn{1}{c}{\textbf{Definition}}           & \multicolumn{1}{c}{\textbf{Examples}}     & \multicolumn{1}{c}{\textbf{F1}} \\  \hline
Start of session            & The tutoring session is just starting. The    & ``Happy to work with you today!'' &  0.79\\
(0.9\%)                     & student and tutor have not yet started a problem. \\ \hline
Start of problem            & The tutor starts a new problem and/or gives   & ``Go ahead and start showing your & 0.70 \\ 
(3.2\%)                     & instructions for the new problem.             & work for this question.''  \\ \hline
During problem attempt      & The student is attempting the problem and/or  & ``Are you working on this problem?'' & 0.70 \\
(48.4\%)                    & the tutor has not yet given away the answer \\
                            & or explanation. \\ \hline
After problem attempt       & The student has attempted the problem and     & ``We know that we cannot subtract & 0.84 \\ 
(40.3\%)                    & the tutor is providing feedback. After a      &  1 - 3, so we will need to borrow \\ 
                            & problem has been attempted, the tutor may want&  from the whole number.'' \\
                            &  to start a new problem (category ``start of \\
                            & problem''). \\\hline
Start of exit ticket        & The tutor starts an exit ticket for the       & ``Now it is time for you to show  & 0.83\\
(1.1\%)                     & student. Note that the exit ticket is a brief & what you have learned by completing \\
                            & assessment near the end of the tutoring       & the Exit Ticket.'' \\
                            & session and the tutor cannot help the student & \\ 
                            & here, unlike for the normal problems.         \\\hline
During exit ticket attempt  & The student is attempting the exit ticket.    & ``I can't help you with the exit  & 0.0 \\ 
                            &                                               & ticket question.''\\ \hline
After exit ticket attempt   & The student has attempted the exit ticket and & Congratulations. You've scored 100\% & 0.90\\
(4.4\%)                     & the tutor is providing feedback. Afterwards,  & in Exit Ticket questions.''\\
                            & the tutor may want to start a new exit ticket & \\ 
                            & (category ``start of exit ticket'').          & \\ \hline   
End of session              & The tutoring session is ending.               & ``We will continue in the next session.'' & 0.98\\ 
(1.8\%)  \\\hline
\end{tabular}
}
\vspace{1em}
\caption{Taxonomy of tutoring moments, including their definitions, examples, and frequency over the labelled dataset of 2,114 messages sent immediately before activation of the Tutor CoPilot tool. 
We train binary classifiers to identify these moments at scale and report their test F1 score as well. 
A majority of Tutor CoPilot usage concetrates during the ``meat'' of student learning: when the student is attempting the problem, or after they have attempted the problem and the tutor is giving them feedback.
The classifier for ``after exit ticket attempt'' scored a low test F1 score, even after tuning the class-imbalance loss, thus we omit its frequency. 
Note that the frequencies do not sum to 1 because the classifiers are not mutually exclusive.
\label{tab:moments}}
\end{table}

\paragraph{Tutor adoption.} 
Which tutors adopted the tool the most?
In Table~\ref{tab:tutor_copilot_adoption}, we report the coefficients on tutor covariates predicting whether the tutor used Tutor CoPilot and number of uses in a given session. 
We also clustered the errors by tutor. 
We found that tutors who have more tutoring experience (measured by how long they've been with the tutoring provider) are less likely to use Tutor CoPilot, and female tutors use Tutor CoPilot one click more than male tutors.


\begin{table}[h!]
\centering
\resizebox{0.5\textwidth}{!}{%
\def\arraystretch{1.15}
\begin{tabular}{lcc}
\toprule
\textbf{Dependent Variable}    & \textbf{Used} & \textbf{Number of Uses }  \\ \hline
\textbf{Gender (Is Female)} & 0.015      & 1.36**  \\
                            & (0.03)    & (0.51) \\
\textbf{Experience (Months)}         & -0.002+    & -0.031+  \\
                            & (0.001)    & (0.02) \\
\textbf{Quality Rating}     & 0.12      & 1.69  \\
                            & (0.08)    & (1.41) \\ \midrule
N                           & 2010      & 2010 \\  \bottomrule

\end{tabular}
}
\vspace{1em}
\caption{Predicting Tutor CoPilot Use From Tutor Characteristics. The parentheses report standard errors clustered by tutor id. $^{+}p<0.1$; $^{*}p<0.05$; $^{**}p<0.01$; $^{***}p<0.001$. \label{tab:tutor_copilot_adoption}}
\end{table}

\section{Student-level analysis \label{app:student_level}}

In this analysis, we measure the effect of treatment exposure on student end-of-year test scores. We instrument treatment exposure as the proportion of the number of tutoring sessions with a treatment tutor over the total number of tutoring sessions attended by the student.
To conduct a student-level analysis, we removed students from our sample who did not participate in any tutoring sessions during the study period. Thus, our analytical sample only includes students with exposure to tutoring, and thus potentially to Tutor CoPilot. 

\begin{figure}[h!]
    \centering
    \includegraphics[width=0.8\linewidth]{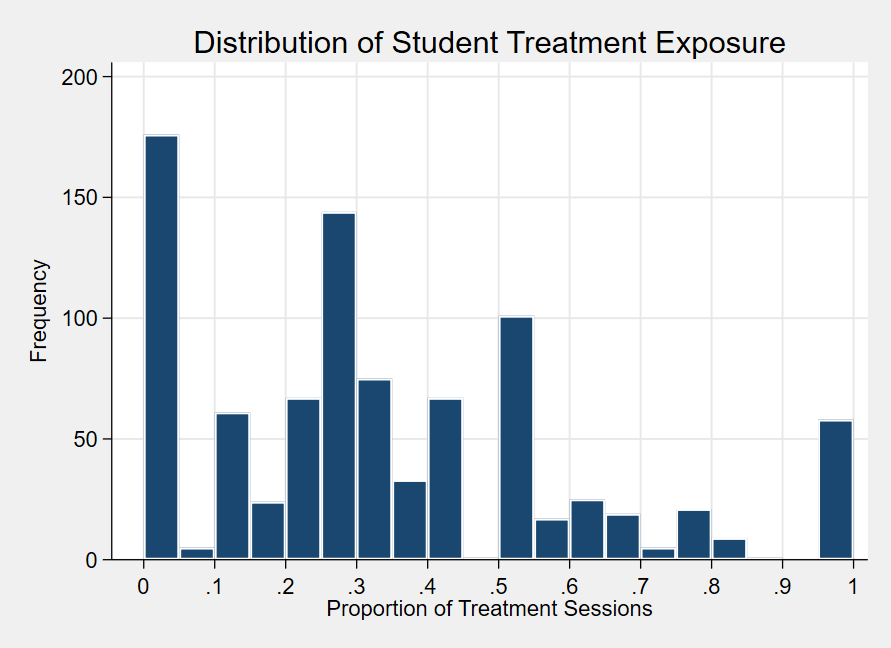}
    \caption{Histogram of student treatment exposure. }
    \label{fig:treatment_exposure}
\end{figure}

\paragraph{Treatment exposure} Figure~\ref{fig:treatment_exposure} reports the histogram on the treatment exposure as a proportion of treatment sessions during the implementation period of Tutor CoPilot for 1,013 students who attended at least one tutoring session with a tutor participating in the study. 
We find that few students had high exposure to Tutor CoPilot, and overall, there was limited variation in treatment exposure across the student population, with most students showing zero or near-zero exposure. 
We therefore expected that the impact on end-of-the-year outcomes would not be significant. 

\paragraph{Regression findings} Table~\ref{tab:student_level_interaction} reports the estimates of the treatment exposure on the student’s EOY MAP Math test scores.
We find no significant effect on the test scores, even after controlling for the number of sessions the student attended during the implementation period of Tutor CoPilot.
The lack of significance may be attributed to the limited variation in treatment exposure among students and insufficient exposure to the treatment, which likely constrained the potential for detecting an effect.


\begin{table}[h!]
\def\sym#1{\ifmmode^{#1}\else\(^{#1}\)\fi}
\centering
\def\arraystretch{1.15}
\resizebox{\textwidth}{!}{%
\begin{tabular}{l*{6}{c}}
\toprule
                    &\multicolumn{3}{c}{\shortstack{Math MAP (std) \\ With Imputation}}&\multicolumn{3}{c}{\shortstack{Math MAP (std)\\ Without Imputation}} \\
\hline
Prop of Treatment Sessions  &     -0.0044         &     -0.0016         &      -0.081         &      -0.022         &      -0.020         &      -0.061         \\
                    &     (0.050)         &     (0.050)         &     (0.082)         &     (0.048)         &     (0.048)         &     (0.078)         \\
[1em]
N Sessions Completed &                     &      0.0097         &      0.0030         &                     &      0.0053         &      0.0017         \\
                    &                     &    (0.0071)         &    (0.0085)         &                     &    (0.0064)         &    (0.0078)         \\
[1em]
&                     &                     &       0.022         &                     &                     &       0.012         \\
                    &                     &                     &     (0.018)         &                     &                     &     (0.017)         \\
\hline
N                   &         895         &         895         &         895         &         853         &         853         &         853         \\
\bottomrule

\end{tabular}
}
\vspace{1em}
\caption{Student-level Estimates of Treatment Exposure Impact on EOY Standardized MAP Math Scores. The sample of students is restricted to students from the partner district who attended at least one tutoring session with a treatment or control tutor during the implementation period for this study. We report the coefficient estimates for models with and without baseline scores (MOY MAP Math) imputations for students missing this information, and the interaction of the percentage of treated sessions and the number of sessions attended during Tutor CoPilot implementation.  Robust standard errors in parentheses. All models include controls for student achievement, characteristics (gender, race/ethnicity, free/reduced lunch, special education, and English learner), and strata (school x grade) fixed effects.  \sym{+} \(p<0.10\), \sym{*} \(p<0.05\), \sym{**} \(p<0.01\), \sym{***} \(p<0.001\).
\label{tab:student_level_interaction}}
\end{table}


\section{Tutor Survey Analysis \label{app:tutor_level}}
The tutor-level analysis is conducted on their end-of-study survey. We report balance checks and differential attrition estimates for the pre- and post-survey respondents in~\ref{app:balance}. We found no significant differences in attrition for survey responses based on treatment assignment.

\paragraph{Post-study survey results}
Table~\ref{tab:tutor_level} reports the treatment estimates on the tutor-level outcomes from the post-study survey. We find no significant changes between the treatment and control groups.


\begin{table}[h!]
\centering
\resizebox{\textwidth}{!}{%
\def\arraystretch{1.15}
\begin{tabular}{lcccc}
\toprule
\textbf{Dependent}      & \textbf{Do you agree or disagree }    & \textbf{How confident are you}    & \textbf{How effective are you at } & \textbf{How much more or less } \\ 
\textbf{Variable}       & \textbf{with this statement: ``My}    & \textbf{at recognizing the kind}  & \textbf{helping students fix their} & \textbf{effective are you at} \\ 
                        & \textbf{students learn more by}       & \textbf{of mathematical}          & \textbf{mistakes?}        & \textbf{helping students fix their} \\ 
                        & \textbf{making mistakes.''?}          & \textbf{mistakes students are}    &                           & \textbf{mistakes now than you } \\ 
                        &                                       & \textbf{making? }                 &                           & \textbf{were three months ago?} \\ \hline
\textbf{Range}          & 1 = Disagree                          & 1 = Not at all confident          & 1 = Not at all effective                           & 1 = A lot less effective \\
                        & 2 = Somewhat disagree                 & 2 = Slightly confident            & 2 = Slightly effective     & 2 = Slightly less effective \\
                        & 3 = Somewhat agree                    & 3 = Confident                     & 3 = Effective             & 3 = No change \\
                        & 4 = Agree                             & 4 = Very Confident                & 4 = Very effective         & 4 = Slightly more effective \\
                        &   &   &   & 5 = A lot more effective \\ \hline
Treatment           &     -0.0058         &       0.078         &       0.023         &      0.0100         \\
                    &     (0.058)         &     (0.057)         &     (0.031)         &     (0.055)         \\
\hline
Control Mean        &       3.472         &       3.350         &       3.811         &       4.601         \\
                    &     (0.040)         &     (0.041)         &     (0.021)         &     (0.040)         \\
Romano-Wolf p-val   &     [0.941]         &     [0.495]         &     [0.851]         &     [0.941]         \\
R-Square            &       0.013         &       0.014         &       0.018         &       0.005         \\
N                   &         726         &         726         &         726         &         726         \\
\bottomrule
\end{tabular}
}
\vspace{1em}
\caption{
Tutor-level Survey Outcomes. The parentheses report robust standard error.  Estimates are from our primary model, which controls for tutor quality rating, months of experience at the platform, and gender. $^{+}p<0.1$; $^{*}p<0.05$; $^{**}p<0.01$; $^{***}p<0.001$. 
\label{tab:tutor_level}}
\end{table}

\section{Compliance \label{app:compliance}}
Out of 2,122 control sessions in our sample, we found six sessions with four different control tutors who had activated the Tutor CoPilot tool. After investigating the issue with the tutoring provider, we discovered that six control tutors were mistakenly given access to Tutor CoPilot on the tutoring platform as if they had been assigned to the treatment condition. We did not obtain information about whether these tutors received training to use the tool or not. Given the negligible number of sessions in which the tool was activated by control tutors, we report all results based on the original assignments.

\section{Alternative Model Specifications\label{app:alternative_specs}}

This section presents alternative specifications to our preferred models for session-level outcomes and strategy use analyses. 

\subsection{Session-level Outcomes}

Table \ref{tab:logit} presents the results for binary session-level outcomes using a logit model.
 Tables \ref{tab:logit_me} and \ref{tab:itt_me} present results for linear and logit models of the session-level outcomes, respectively, using random effects for students and tutors separately. These estimates reveal that our main results remain significant across specifications, despite variations in significance. In particular, the treatment effect for exit tickets passed conditional on attempt is estimated as significant at a 5\% or 10\% level, and the treatment effect for unconditional exit tickets is estimated as significant at a 1\% or 5\% level depending on the model specification.

We also present estimates for alternative specifications for the tutor quality and experience heterogeneity analyses in Table ~\ref{tab:heterog_qs_age}.

\begin{table}[h!]\centering
\def\sym#1{\ifmmode^{#1}\else\(^{#1}\)\fi}
\resizebox{0.6\textwidth}{!}{
\begin{tabular}{l*{3}{c}}
\toprule
                    &\multicolumn{1}{c}{\shortstack{Exit Tickets \\ Attempted}}&\multicolumn{1}{c}{\shortstack{Exit Tickets \\ Passed \\ Conditional}}&\multicolumn{1}{c}{\shortstack{Exit Tickets \\ Passed \\ Unconditional}}\\
\midrule
Treatment           &        0.17\sym{+}  &        0.18\sym{*}  &        0.19\sym{**} \\
                    &     (0.093)         &     (0.080)         &     (0.069)         \\
\hline
Romano-Wolf p-val   &     [0.030]         &     [0.030]         &     [0.010]         \\
Z                   &       1.803         &       2.202         &       2.769         \\
Odds Ratio          &       1.182         &       1.193         &       1.209         \\
N                   &        4105         &        3518         &        4133         \\
\bottomrule

\end{tabular}
}
\vspace{1em}
\caption{ITT Logit Model Results for Binary Outcomes. Controls for baseline math scores, student demographics, and fixed effects for strata (school $\times$ grade). Standard errors clustered at the tutor-student pair are shown in parentheses, and Romano-Wolf p-values adjusted for multiple hypotheses are shown in brackets.  \label{tab:logit}}
\end{table}


\begin{table}[h!]\centering
\def\sym#1{\ifmmode^{#1}\else\(^{#1}\)\fi}
\resizebox{0.6\textwidth}{!}{
\begin{tabular}{l*{3}{c}}
\toprule
                    &\multicolumn{1}{c}{\shortstack{Exit Tickets \\ Attempted}}&\multicolumn{1}{c}{\shortstack{Exit Tickets \\ Passed \\ Conditional}}&\multicolumn{1}{c}{\shortstack{Exit Tickets \\ Passed \\ Unconditional}}\\
\hline
Treatment           &        0.18         &        0.16\sym{+}  &        0.21\sym{*}  \\
                    &      (0.14)         &     (0.093)         &     (0.096)         \\
\hline
Z                   &       1.231         &       1.769         &       2.203         \\
Odds Ratio          &       1.195         &       1.179         &       1.236         \\
N                   &        4105         &        3518         &        4133         \\
\bottomrule
\multicolumn{4}{l}{\footnotesize }\\
\end{tabular}
}
\vspace{1em}
\caption{ITT Logit Model Results for Binary Outcomes with Random Effects for Students and Tutors. Controls for baseline math scores, student demographics, and fixed effects for strata (school $\times$ grade). Standard errors shown in parentheses.  \label{tab:logit_me}}
\end{table}



\begin{table}[h!]
\centering
\resizebox{\textwidth}{!}{
\def\arraystretch{1.15}
\begin{tabular}{lccccc}
\toprule
\textbf{} & \multicolumn{5}{c}{\textbf{Panel A. Session outcomes}} \\
                    &\multicolumn{1}{c}{\shortstack{\textbf{Participation} \\ \textbf{Points} \\ \textcolor{white}{.} } }&\multicolumn{1}{c}{\shortstack{\textbf{Participation} \\ \textbf{Points} \\ \textbf{Standardized}}}&\multicolumn{1}{c}{\shortstack{ \textbf{Exit Tickets} \\ \textbf{Attempted} \\ \textcolor{white}{.} }}&\multicolumn{1}{c}{\shortstack{\textbf{Exit Tickets} \\ \textbf{Passed} \\ \textbf{Conditional}}}&\multicolumn{1}{c}{\shortstack{ \textbf{Exit Tickets} \\ \textbf{Passed} \\ \textbf{Unconditional}}}\\
\hline
Treatment           &      -0.022         &     -0.0013         &       0.017         &       0.028\sym{+}  &       0.037\sym{*}  \\
                    &      (0.33)         &     (0.034)         &     (0.013)         &     (0.016)         &     (0.016)         \\
\hline
Control Mean        &      14.035         &       0.012         &       0.842         &       0.733         &       0.617         \\
                    &     (0.234)         &     (0.024)         &     (0.009)         &     (0.011)         &     (0.011)         \\
N                   &    4136         &    4136         &    4136         &    3521        &    4136         \\
\hline
\textbf{} & \multicolumn{5}{c}{\textbf{Panel B. Student survey outcomes}} \\ 

                    &\multicolumn{1}{c}{\shortstack{\textbf{My Tutor cared about} \\ \textbf{understanding} \\ \textbf{math over} \\ \textbf{memorizing} \\ \textbf{the solution.}}}&\multicolumn{1}{c}{\shortstack{ \textbf{My tutor cared} \\ \textbf{about how well I} \\ \textbf{do in math.} \\ \textcolor{white}{.}}}&\multicolumn{1}{c}{\shortstack{ \textbf{Even when math} \\ \textbf{is hard, I know} \\ \textbf{I can learn it.} \\ \textcolor{white}{.} }}&\multicolumn{1}{c}{\shortstack{ \textbf{Session} \\ \textbf{Rating} \\ \textcolor{white}{.} }}&\multicolumn{1}{c}{\shortstack{ \textbf{Tutor} \\ \textbf{Rating} \\ \textcolor{white}{.} }} \\
\hline
Treatment           &      0.0010         &       0.029         &       0.020         &      0.0041         &       0.041         \\
                    &     (0.057)         &     (0.053)         &     (0.056)         &     (0.037)         &     (0.039)         \\
\hline
Control Mean        &       4.180         &       4.303         &       4.238         &       4.760         &       4.734         \\
                    &     (0.039)         &     (0.036)         &     (0.038)         &     (0.025)         &     (0.027)         \\
N                   &    1931         &    1931         &    1931         &    1948       &    1952         \\
\bottomrule
\end{tabular}
}
\vspace{1em}
\caption{ITT Linear Model Results for Student Session-level Outcomes with Random Effects for Students and Tutors. The parentheses report the standard error. Models include controls for baseline math scores, student demographics, and a fixed effect for strata (school $\times$ grade). Participation points are standardized within-sample, by grade. The survey items are on a 5-point scale where higher is better. $^{+}p<0.1$; $^{*}p<0.05$; $^{**}p<0.01$; $^{***}p<0.001$. \label{tab:itt_me}}
\end{table}

\begin{table}[h!]\centering
\def\sym#1{\ifmmode^{#1}\else\(^{#1}\)\fi}
\resizebox{0.8\textwidth}{!}{
\begin{tabular}{l*{3}{c}}
\toprule
                    &\multicolumn{1}{c}{\shortstack{Exit Tickets \\ Passed \\ Unconditional}}&\multicolumn{1}{c}{\shortstack{Exit Tickets \\ Passed \\ Unconditional}}&\multicolumn{1}{c}{\shortstack{Exit Tickets \\ Passed \\ Unconditional}}\\
\hline
Treatment           &      0.0869\sym{+}  &       0.204\sym{*}  &     0.00184         \\
                    &    (0.0449)         &    (0.0965)         &    (0.0346)         \\
[1em]
Treat x Tutor Quality Rating (Cont)&     -0.0891         &      -0.392         &                     \\
                    &    (0.0747)         &     (0.299)         &                     \\
[1em]
Treat x Tutor Experience (Cont)&   -0.000395         &    -0.00746         &                     \\
                    &   (0.00123)         &   (0.00924)         &                     \\
[1em]
Treat x Tutor Quality Rating Sq&                     &       0.494         &                     \\
                    &                     &     (0.752)         &                     \\
[1em]
Treat x Tutor Experience Sq&                     &    0.000135         &                     \\
                    &                     &  (0.000244)         &                     \\
[1em]
Treat x Tutor QR x QR Sq&                     &      -0.165         &                     \\
                    &                     &     (0.574)         &                     \\
[1em]
Treat x Tutor Exp x Exp Sq&                     &   -7.81e-09         &                     \\
                    &                     &  (3.54e-08)         &                     \\
[1em]
Treat x Tutor Quality Rating Low&                     &                     &      0.0587         \\
                    &                     &                     &    (0.0361)         \\
[1em]
Treat x Tutor Quality Rating Medium&                     &                     &     0.00298         \\
                    &                     &                     &    (0.0361)         \\
[1em]
Treat x Tutor Experience Low&                     &                     &      0.0587         \\
                    &                     &                     &    (0.0363)         \\
[1em]
Treat x Tutor Experience Medium&                     &                     &    -0.00723         \\
                    &                     &                     &    (0.0371)         \\
\hline
N                   &        4136         &        4136         &        4136         \\
\bottomrule
\vspace{.5em}
\end{tabular}
}
\caption{Regression Results from the Tutor Quality and Experience Heterogeneity Analyses. Column 1 models the interaction between the treatment and tutor quality and experience linear measures, column 2 includes a quadratic term and interactions for these measures, and column 3 shows the results for a model that categorizes the level of quality and experience into "Low", "Medium", and "High", based on tercile grouping. Figure 2 is based on the latter. Estimates include controls for baseline math scores, student demographics, and fixed effects for strata (school $\times$ grade). Standard errors clustered for student-tutor pairs in parentheses. \sym{+} \(p<0.10\), \sym{*} \(p<0.05\), \sym{**} \(p<0.01\), \sym{***} \(p<0.001\) \label{tab:heterog_qs_age}}
\end{table}


\subsection{Strategy Use Analyses  \label{app:tutor_logodds}}

We present the results of the log odds analysis of the strategies used by tutors considering different specifications. Table~\ref{tab:strategies_clustered} in the manuscript presents the results for our preferred model, described in the Methods section, which includes student-tutor pair clustering to account for repeated interactions between students and tutors, following our pre-registered specification for session-level outcomes. 
We find the estimates for ``prompting the student to explain'' and ``asking questions to guide thinking'' remain significant and in favor of treatment tutors becoming more likely to use expert-recommended practices. However, most of the other strategies investigated are not significantly different between treatment and control tutors under this specification, with the exception of ``encourage student in a generic way'', which is more likely to be used by control tutors.

\begin{figure}[h!]
    \centering
    \begin{subfigure}[b]{0.95\textwidth}
        \centering
        \includegraphics[width=\textwidth]{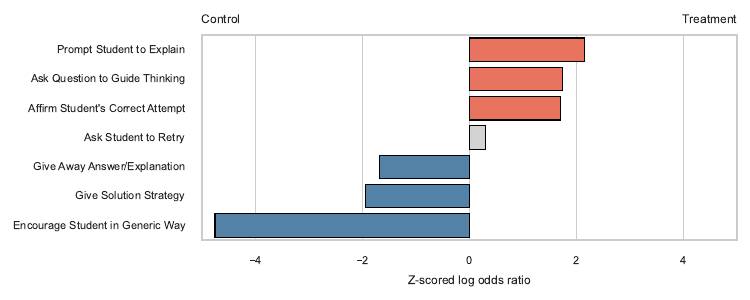}
    \end{subfigure}
    \caption{Log odds Analysis as proposed in Monroe et al.~(2008)\cite{monroe2008fightin}. Strategies More Likely to be Used by Control Tutors (left) vs. Treatment Tutors (right). 
    Strategies with a z-score below 1 standard deviation are shaded in gray.   
    \label{fig:strategies}}
\end{figure}

Our first alternative specification adapted the Fightin’ Words method from Monroe et al.~(2008)~\cite{monroe2008fightin}, which calculated the z-score log odds to compare the frequency of strategies while adjusting for a prior distribution. This method quantifies the causal differences in strategies between treatment and control tutors, but does not account for the multi-level structure of our data. We report the z-scored log odds for this method in Figure~\ref{fig:strategies}. Without accounting for the correlation from repeated interactions between same student-tutor pairs and the multi-level structure of the data, we found that treatment tutors were more likely to use strategies such as ``prompting the student to explain'' and ``asking questions to guide thinking'', which align with expert-recommended practices for promoting deeper learning \cite{chi1989selfexplanation,lepper2002wisdom}. In contrast, control tutors appear to rely on strategies that focus on directly leading students to the solution or providing passive support. 
These strategies included immediately giving away the answer or providing the solution to that specific problem, which help students complete tasks but do less to develop their deeper understanding.


\begin{table}[h!]
\centering
\resizebox{\textwidth}{!}{
\def\arraystretch{1.15}
\begin{tabular}{l*{7}{c}}
\toprule
 & \shortstack{ Prompt \\ Student \\ to Explain} & \shortstack{Ask \\ Question \\ to Guide \\ Thinking} & \shortstack{Affirm \\ Student's \\ Correct \\ Attempt} & \shortstack{Ask \\ Student \\ to Retry} & \shortstack{Give \\ Away \\ Answer/ \\ Explanation} & \shortstack{ Give \\ Solution \\ Strategy} & \shortstack{Encourage \\ Student in \\ Generic \\ Way}    \\
\midrule
Treatment           &        0.12\sym{*}  &       0.039         &       0.038         &        0.12         &      -0.039         &     -0.0056         &       -0.10\sym{*}  \\
                    &     (0.057)         &     (0.045)         &     (0.031)         &      (0.10)         &     (0.043)         &     (0.044)         &     (0.043)         \\
\hline
Z                   &       2.107         &       0.861         &       1.250         &       1.176         &      -0.900         &      -0.128         &      -2.416         \\
Odds Ratio          &       1.127         &       1.040         &       1.039         &       1.126         &       0.962         &       0.994         &       0.901         \\
N                   &      241005         &      241005         &      241005         &      241005         &      241005         &      241005         &      241005         \\

\bottomrule
\end{tabular}
}
\vspace{1em}
\caption{ Alternative Specification - Log Odds Analysis of Strategy Use Differences Between Treatment and Control Tutors. This table reports the estimates from a logit model with random effects for tutors and students separately.  The parentheses report the standard error.  $^{+}p<0.1$; $^{*}p<0.05$; $^{**}p<0.01$; $^{***}p<0.001$. \label{tab:strategies_melogit}}
\end{table}


The second alternative specification we present here uses a logit specification with random effects for students and tutors separately as an alternative to student-tutor pair clustering to account for the multi-level structure of our data. We report the results for this specification in Table~\ref{tab:strategies_melogit}. Based on these estimates, differences between treatment and control tutors for two strategies were found to be significant and robust across specifications. Despite the estimated difference for ``asking questions to guide thinking'' not being significant under this model, we still find treatment tutors to be more likely to ``prompt the student to explain'' than control tutors, and less likely to ``encourage student in a generic way''.

\paragraph{Heterogeneity Analysis.} Tables \ref{tab:strategies_qs_Low},
\ref{tab:strategies_qs_Medium},
\ref{tab:strategies_qs_High}, 
\ref{tab:strategies_exp_Low}, \ref{tab:strategies_exp_Medium}, and
\ref{tab:strategies_exp_High},  show the results of the log-odds analysis of the use of different types of remediation strategies separately for tutors grouped into quality rating and experience terciles (low, medium, high). Quality rating refers to tutor quality score determined prior to the study by the tutoring provider based on session observations and quality rubric scores, and experience is defined as the amount of time tutors had been employed as a tutor at this tutoring provider.   

Based on the results for tutor groups categorized as low, medium, and high quality score terciles (Tables \ref{tab:strategies_qs_Low},
\ref{tab:strategies_qs_Medium},
\ref{tab:strategies_qs_High}, respectively), we found the most significant change for low quality tutors in the treatment to be a higher likelihood to use multiple of the strategies evaluated, in particular two considered beneficial to promote deeper learning. We also found differences between treatment and control tutors in the medium and high quality groups. Most notably, treatment tutors in both quality groups were more likely to ``ask a question to guide thinking'' than control tutors. However, the groups differed in the strategies treatment tutors used less than their control counterparts -- ``encouraged students in a generic way'' for medium quality tutors, and ``gave away the answer/explanation'' for high quality tutors in the treatment group. Both of which are considered lower quality strategies based on the taxonomy used to develop our strategy classifiers (Table~\ref{tab:strategies}).

Our results for tutors of different experience level are shown in Tables \ref{tab:strategies_exp_Low},
\ref{tab:strategies_exp_Medium},
\ref{tab:strategies_exp_High} for low, medium, and high terciles, respectively. We found that less experienced tutors on the platform who were assigned to the treatment were less likely to ``encourage student in a generic way'' but did not become more likely to use any of the strategies evaluated. Treatment tutors categorized as medium experienced had results consistent with our main findings, displaying a higher likelihood of ``prompting students to explain'' and lower likelihood of ``encouraging students in a generic way'' compared to their counterparts in the control group. For the most experienced tutors in our sample, the treatment group was found to be less likely to ``give away answers or explanations''.

Tutors at different levels of quality or experience were not impacted in the same way by the tool. However, these estimates show positive evidence of the tool on strategy use, such as reducing the use of lower quality strategies or increasing the use of higher quality ones.

\begin{table}[h!]
\centering
\resizebox{\textwidth}{!}{
\def\arraystretch{1.15}
\begin{tabular}{l*{7}{c}}
\toprule
  & \shortstack{ Prompt \\ Student \\ to Explain} & \shortstack{Ask \\ Question \\ to Guide \\ Thinking} & \shortstack{Affirm \\ Student's \\ Correct \\ Attempt} & \shortstack{Ask \\ Student \\ to Retry} & \shortstack{Give \\ Away \\ Answer/ \\ Explanation} & \shortstack{ Give \\ Solution \\ Strategy} & \shortstack{Encourage \\ Student in \\ Generic \\ Way}    \\
\midrule
Treatment           &        0.19\sym{**} &      -0.087         &        0.12\sym{**} &        0.18\sym{+}  &     0.00095         &    -0.00045         &      -0.058         \\
                    &     (0.061)         &     (0.053)         &     (0.038)         &      (0.10)         &     (0.052)         &     (0.049)         &     (0.045)         \\
\hline
Romano-Wolf p-val   &     [0.010]         &     [0.356]         &     [0.010]         &     [0.228]         &     [1.000]         &     [1.000]         &     [0.723]         \\
Z                   &       3.199         &      -1.628         &       3.063         &       1.791         &       0.018         &      -0.009         &      -1.293         \\
Odds Ratio          &       1.214         &       0.917         &       1.124         &       1.198         &       1.001         &       1.000         &       0.944         \\
N                   &       82952         &       82952         &       82952         &       82952         &       82952         &       82952         &       82952         \\
\bottomrule
\end{tabular}
}
\vspace{1em}
\caption{ Low Quality Rating Score Tutors - Log odds analysis of strategy use. The parentheses report the standard error
clustered by student-tutor pairs. Romano-Wolf adjusted p-values are shown in brackets.  $^{+}p<0.1$; $^{*}p<0.05$; $^{**}p<0.01$; $^{***}p<0.001$. \label{tab:strategies_qs_Low}}
\end{table}


\begin{table}[h!]
\centering
\resizebox{\textwidth}{!}{
\def\arraystretch{1.15}
\begin{tabular}{l*{7}{c}}
\toprule
   & \shortstack{ Prompt \\ Student \\ to Explain} & \shortstack{Ask \\ Question \\ to Guide \\ Thinking} & \shortstack{Affirm \\ Student's \\ Correct \\ Attempt} & \shortstack{Ask \\ Student \\ to Retry} & \shortstack{Give \\ Away \\ Answer/ \\ Explanation} & \shortstack{ Give \\ Solution \\ Strategy} & \shortstack{Encourage \\ Student in \\ Generic \\ Way}    \\
\midrule
Treatment           &        0.10\sym{+}  &        0.12\sym{*}  &      -0.025         &       -0.12         &      -0.034         &       0.036         &       -0.26\sym{***}\\
                    &     (0.063)         &     (0.054)         &     (0.038)         &      (0.11)         &     (0.051)         &     (0.053)         &     (0.047)         \\
\hline
Romano-Wolf p-val   &     [0.317]         &     [0.020]         &     [1.000]         &     [0.911]         &     [1.000]         &     [1.000]         &     [0.010]         \\
Z                   &       1.675         &       2.284         &      -0.667         &      -1.034         &      -0.678         &       0.677         &      -5.594         \\
Odds Ratio          &       1.110         &       1.131         &       0.975         &       0.889         &       0.966         &       1.037         &       0.770         \\
N                   &       78358         &       78358         &       78358         &       78358         &       78358         &       78358         &       78358         \\
\bottomrule
\end{tabular}
}
\vspace{1em}
\caption{ Medium Quality Rating Score Tutors - Log odds analysis of strategy use. The parentheses report the standard error
clustered by student-tutor pairs. Romano-Wolf adjusted p-values are shown in brackets.  $^{+}p<0.1$; $^{*}p<0.05$; $^{**}p<0.01$; $^{***}p<0.001$. \label{tab:strategies_qs_Medium}}
\end{table}


\begin{table}[h!]
\centering
\resizebox{\textwidth}{!}{
\def\arraystretch{1.15}
\begin{tabular}{l*{7}{c}}
\toprule
 & \shortstack{ Prompt \\ Student \\ to Explain} & \shortstack{Ask \\ Question \\ to Guide \\ Thinking} & \shortstack{Affirm \\ Student's \\ Correct \\ Attempt} & \shortstack{Ask \\ Student \\ to Retry} & \shortstack{Give \\ Away \\ Answer/ \\ Explanation} & \shortstack{ Give \\ Solution \\ Strategy} & \shortstack{Encourage \\ Student in \\ Generic \\ Way}    \\
\midrule
Treatment           &      -0.015         &        0.14\sym{**} &       0.012         &      -0.057         &       -0.11\sym{*}  &      -0.039         &       0.029         \\
                    &     (0.065)         &     (0.053)         &     (0.038)         &      (0.11)         &     (0.055)         &     (0.053)         &     (0.042)         \\
\hline
Romano-Wolf p-val   &     [1.000]         &     [0.010]         &     [1.000]         &     [1.000]         &     [0.050]         &     [0.990]         &     [1.000]         \\
Z                   &      -0.236         &       2.740         &       0.316         &      -0.493         &      -2.018         &      -0.749         &       0.697         \\
Odds Ratio          &       0.985         &       1.156         &       1.012         &       0.945         &       0.895         &       0.961         &       1.030         \\
N                   &       79695         &       79695         &       79695         &       79695         &       79695         &       79695         &       79695         \\
\bottomrule
\end{tabular}
}
\vspace{1em}
\caption{ High Quality Rating Score Tutors - Log odds analysis of strategy use. The parentheses report the standard error
clustered by student-tutor pairs. Romano-Wolf adjusted p-values are shown in brackets.  $^{+}p<0.1$; $^{*}p<0.05$; $^{**}p<0.01$; $^{***}p<0.001$. \label{tab:strategies_qs_High}}
\end{table}


\begin{table}[h!]
\centering
\resizebox{\textwidth}{!}{
\def\arraystretch{1.15}
\begin{tabular}{l*{7}{c}}
\toprule
   & \shortstack{ Prompt \\ Student \\ to Explain} & \shortstack{Ask \\ Question \\ to Guide \\ Thinking} & \shortstack{Affirm \\ Student's \\ Correct \\ Attempt} & \shortstack{Ask \\ Student \\ to Retry} & \shortstack{Give \\ Away \\ Answer/ \\ Explanation} & \shortstack{ Give \\ Solution \\ Strategy} & \shortstack{Encourage \\ Student in \\ Generic \\ Way}    \\
\midrule
Treatment           &      -0.027         &       0.082         &       0.036         &       0.065         &      -0.010         &      0.0023         &       -0.13\sym{**} \\
                    &     (0.063)         &     (0.053)         &     (0.036)         &      (0.11)         &     (0.050)         &     (0.049)         &     (0.041)         \\
\hline
Romano-Wolf p-val   &     [1.000]         &     [0.406]         &     [0.931]         &     [1.000]         &     [1.000]         &     [1.000]         &     [0.010]         \\
Z                   &      -0.436         &       1.565         &       0.980         &       0.578         &      -0.202         &       0.047         &      -3.079         \\
Odds Ratio          &       0.973         &       1.086         &       1.036         &       1.067         &       0.990         &       1.002         &       0.882         \\
N                   &       88684         &       88684         &       88684         &       88684         &       88684         &       88684         &       88684         \\
\bottomrule
\end{tabular}
}
\vspace{1em}
\caption{ Low Experience Tutors - Log odds analysis of strategy use. The parentheses report the standard error
clustered by student-tutor pairs. Romano-Wolf adjusted p-values are shown in brackets.  $^{+}p<0.1$; $^{*}p<0.05$; $^{**}p<0.01$; $^{***}p<0.001$. \label{tab:strategies_exp_Low}}
\end{table}


\begin{table}[h!]
\centering
\resizebox{\textwidth}{!}{
\def\arraystretch{1.15}
\begin{tabular}{l*{7}{c}}
\toprule
  & \shortstack{ Prompt \\ Student \\ to Explain} & \shortstack{Ask \\ Question \\ to Guide \\ Thinking} & \shortstack{Affirm \\ Student's \\ Correct \\ Attempt} & \shortstack{Ask \\ Student \\ to Retry} & \shortstack{Give \\ Away \\ Answer/ \\ Explanation} & \shortstack{ Give \\ Solution \\ Strategy} & \shortstack{Encourage \\ Student in \\ Generic \\ Way}    \\
\midrule
Treatment           &        0.29\sym{***}&       0.011         &       0.048         &       -0.19\sym{+}  &       0.021         &      -0.012         &       -0.16\sym{***}\\
                    &     (0.062)         &     (0.055)         &     (0.038)         &      (0.11)         &     (0.055)         &     (0.052)         &     (0.046)         \\
\hline
Romano-Wolf p-val   &     [0.010]         &     [1.000]         &     [0.723]         &     [0.228]         &     [1.000]         &     [1.000]         &     [0.010]         \\
Z                   &       4.636         &       0.208         &       1.271         &      -1.791         &       0.389         &      -0.221         &      -3.558         \\
Odds Ratio          &       1.331         &       1.012         &       1.049         &       0.827         &       1.022         &       0.988         &       0.851         \\
N                   &       78381         &       78381         &       78381         &       78381         &       78381         &       78381         &       78381         \\
\bottomrule
\end{tabular}
}
\vspace{1em}
\caption{ Medium Experience Tutors - Log odds analysis of strategy use. The parentheses report the standard error
clustered by student-tutor pairs. Romano-Wolf adjusted p-values are shown in brackets.  $^{+}p<0.1$; $^{*}p<0.05$; $^{**}p<0.01$; $^{***}p<0.001$. \label{tab:strategies_exp_Medium}}
\end{table}



\begin{table}[h!]
\centering
\resizebox{\textwidth}{!}{
\def\arraystretch{1.15}
\begin{tabular}{l*{7}{c}}
\toprule
 & \shortstack{ Prompt \\ Student \\ to Explain} & \shortstack{Ask \\ Question \\ to Guide \\ Thinking} & \shortstack{Affirm \\ Student's \\ Correct \\ Attempt} & \shortstack{Ask \\ Student \\ to Retry} & \shortstack{Give \\ Away \\ Answer/ \\ Explanation} & \shortstack{ Give \\ Solution \\ Strategy} & \shortstack{Encourage \\ Student in \\ Generic \\ Way}    \\
\midrule
Treatment           &       0.029         &       0.077         &     0.00064         &        0.20\sym{+}  &       -0.14\sym{**} &       0.032         &      0.0052         \\
                    &     (0.064)         &     (0.053)         &     (0.040)         &      (0.11)         &     (0.054)         &     (0.053)         &     (0.047)         \\
\hline
Romano-Wolf p-val   &     [1.000]         &     [0.475]         &     [1.000]         &     [0.228]         &     [0.010]         &     [1.000]         &     [1.000]         \\
Z                   &       0.454         &       1.464         &       0.016         &       1.795         &      -2.587         &       0.610         &       0.110         \\
Odds Ratio          &       1.029         &       1.081         &       1.001         &       1.216         &       0.870         &       1.033         &       1.005         \\
N                   &       73940         &       73940         &       73940         &       73940         &       73940         &       73940         &       73940         \\
\bottomrule
\end{tabular}
}
\vspace{1em}
\caption{ High Experience Tutors - Log odds analysis of strategy use. The parentheses report the standard error
clustered by student-tutor pairs. Romano-Wolf adjusted p-values are shown in brackets.  $^{+}p<0.1$; $^{*}p<0.05$; $^{**}p<0.01$; $^{***}p<0.001$. \label{tab:strategies_exp_High}}
\end{table}


\section{Treatment on the Treated Analysis \label{app:tot}}

\paragraph{Setup.}
In the main analysis, we estimate the intention-to-treat (ITT) effect, the impact of being assigned to the treatment group, using linear regression. However, treatment assignment alone may not fully capture the effect of the intervention.
Tutors in some sessions may not use Tutor CoPilot, while in other sessions they may use it. To disentangle the effect of merely having access to the tool from the effect of actually using it, we extend our analysis to a treatment on the treated analysis (TOT) using a two-stage least squares (2SLS) regression approach. This framework helps isolate the causal effect of tool usage, accounting for the possibility that tool use is endogenous (e.g., tutors may be more likely to use the tool in certain kinds of sessions, such as when the student makes a mistake).

Intuitively, the 2SLS regression works as follows: In the first stage, we predict the likelihood of a tutor using Tutor CoPilot during a session based on their treatment assignment. This step isolates the portion of tool usage that is plausibly exogenous — that is, due to the random assignment rather than other tutor- or session-specific factors. In the second stage, we use the predicted probability of tool use from the first stage as an explanatory variable to estimate the impact on the outcomes of interest. 
This two-stage procedure yields an estimate of the local average treatment effect (LATE) for tutors whose usage behavior was influenced by the treatment assignment, also known as the treatment-on-the-treated (ToT) effect. This method helps us understand the causal effect of using Tutor CoPilot, beyond just having access to it.

Concretely, we extend our ITT model to the 2SLS framework to estimate the ToT effect by breaking the estimation process into two stages and incorporating a variable that indicates whether the tutor used the tool in the session or not into the estimation process. In the first stage, we use the treatment assignment, $\text{Treatment}_{t}$, and the covariates from our ITT model (student demographics and school-grade fixed effects) to predict whether the tutor used Tutor CoPilot, $\text{Used}_{ist}$. This predicted use is then used in the second stage to predict the outcomes. We include the full list of covariates used in the ITT in both estimation stages and cluster standard errors by student-tutor pairs.

\begin{align}
    \text{Used}_{ist}= \pi_0 + \pi_1 \text{Treatment}_{t} + \delta X_s + \omega_{k(s)} + \nu_{ist} \\
    Y_{ist}= \beta_0 + \beta_1 \hat{\text{Used}}_{ist} + \lambda X_s + \omega_{k(s)} + \zeta_{ist} 
\end{align}

\paragraph{Findings.} We report the TOT findings in Table~\ref{tab:tot}. We find a much larger, significant effect of treatment on our main outcome variables: Students are 14 p.p. more likely to pass their exit tickets when tutor use Tutor CoPilot (p<0.01), resulting in a 62\% $\rightarrow$ 76\% passing rate.


\begin{table}[h!]
\centering
\resizebox{\textwidth}{!}{%
\def\arraystretch{1.15}
\begin{tabular}{lccccc}
\toprule
\textbf{} & \multicolumn{5}{c}{\textbf{Panel A. Session outcomes}} \\
\textbf{Dependent}    & \textbf{Participation} & \textbf{Participation }  & \textbf{Exit Ticket} & \textbf{Exit Ticket} & \textbf{Exit Ticket} \\ 
\textbf{ Variable:}   & \textbf{Points} & \textbf{Points}  & \textbf{Attempted} & \textbf{Passed} & \textbf{Passed} \\ 
                                & \textbf{} & \textbf{ (Standardized)}  & \textbf{} & \textbf{(Conditional)} & \textbf{ (Unconditional)} \\ 
\hline
\textbf{Treatment Effect}   &        0.32         &       0.035         &       0.064\sym{+}  &        0.10\sym{*}  &        0.14\sym{**} \\
                    &      (0.93)         &     (0.095)         &     (0.037)         &     (0.047)         &     (0.050)         \\
\hline
Romano-Wolf p-val   &     [0.990]         &     [0.990]         &     [0.129]         &     [0.040]         &     [0.010]         \\
R-Square            &       0.083         &       0.085         &       0.019         &       0.029         &       0.040         \\
N                   &        4136         &        4136         &        4136         &        3521         &        4136         \\
\hline
\textbf{} & \multicolumn{5}{c}{\textbf{Panel B. Student survey outcomes}} \\ 
\textbf{Dependent}      & \textbf{My tutor cared} & \textbf{My tutor cared} & \textbf{Even when math} & \textbf{Session Rating} & \textbf{Tutor Rating} \\ 
\textbf{Variable:}      & \textbf{understanding} & \textbf{about how well I} & \textbf{is hard, I know I} & \textbf{} & \textbf{} \\ 
\textbf{ }              & \textbf{math over} & \textbf{do in math.} & \textbf{can learn it.} & \textbf{} & \textbf{} \\ 
\textbf{ }              & \textbf{memorizing} & \textbf{} & \textbf{} & \textbf{} & \textbf{} \\ 
\textbf{}               & \textbf{ the solution.} & \textbf{} & \textbf{} & \textbf{} & \textbf{} \\ 
\hline
\textbf{Treatment Effect}   &      -0.013         &       0.086         &       0.059         &     -0.0054         &       0.093         \\
                    &      (0.19)         &      (0.18)         &      (0.19)         &      (0.13)         &      (0.13)         \\
\hline
Romano-Wolf p-val   &     [1.000]         &     [0.950]         &     [1.000]         &     [1.000]         &     [0.812]         \\
R-Square            &       0.022         &       0.035         &       0.032         &       0.021         &       0.022         \\
N                   &        1931         &        1931         &        1931         &        1948         &        1952         \\
\bottomrule
\end{tabular}
}
\vspace{1em}
\caption{TOT Analysis on Student Session-level Outcomes. The parentheses report the standard error clustered by student-tutor pairs. Estimates are from the second stage of the 2SLS model, which includes controls for baseline math scores, student demographics, and fixed effects for strata (school $\times$ grade). Participation points are standardized within-sample and by grade. The survey items are on a 5-point scale where higher is better. Romano-Wolf adjusted p-values are shown in brackets. $^{+}p<0.1$; $^{*}p<0.05$; $^{**}p<0.01$; $^{***}p<0.001$.  \label{tab:tot}}
\end{table}

\section{Structured Interview Protocol with Tutors  \label{app:tutor_interview}}
To understand how tutors perceive the Tutor CoPilot tool, we conducted an interview following a structured interview protocol a week after the study had concluded. The interview was done virtually. 
One of the authors of this work was the discussion lead, and the discussion included 18 tutors, alongside members of the tutoring operations team and an engineering manager. 
The structured format was chosen to facilitate more focused and productive discussions, as it can be challenging for tutors to provide feedback in an unstructured group setting.

The interviews were conducted in a 1-hour session, following a specific protocol:
\begin{enumerate}
    \item \textbf{Session Timing}: The session was scheduled to last one hour. We allowed the first five minutes for everyone to join, with the discussion beginning promptly at :05.
    \item \textbf{Recording Consent}: At :05, the discussion lead asked participants if they were comfortable with the session being recorded on Zoom. Once consent was obtained, the session was recorded to ensure accurate capture of the feedback provided.
    \item \textbf{Session Introduction}: The discussion lead communicated that the purpose of the session was to gather their feedback on Tutor CoPilot. The lead emphasized that the session was not an evaluation of the tutors, but rather an opportunity for the tutors to share their experiences and thoughts on Tutor CoPilot. This would inform how future iterations of Tutor CoPilot could be improved.
    \item \textbf{Structured Questions}: Participants were provided with a list of questions in a shared Google document. The questions were:
    \begin{itemize}
        \item Have you used Tutor CoPilot? If so, how?
        \item Can you provide an example of how you’ve used it?
        \item How has your use of Tutor CoPilot changed over time?
        \item What aspects of Tutor CoPilot do you like? What do you not like?
        \item What would you like Tutor CoPilot to do for you? Are there specific times during tutoring sessions when you would like additional support from the CoPilot?
        \item Do you have any other thoughts or feedback on Tutor CoPilot?
    \end{itemize}
    \item \textbf{Reflection Period}: Participants were given seven minutes, ending at :20, to reflect on the questions and note down their responses in the shared document.
    \item \textbf{Group Discussion}: Following the reflection period, we went through the responses together, allowing for a discussion of the feedback shared. This part of the session was intended to foster a collaborative exchange of ideas and experiences among the tutors.
    \item \textbf{Conclusion}: The session was concluded by the researcher thanking participants for their input. We also encouraged participants to send additional thoughts or questions to the researcher's provided email address.
\end{enumerate}

\section{Balance Checks and Attrition \label{app:balance}}

We conducted balance checks to ensure that the tutor sample is comparable across treatment and control groups in terms of key covariates.  
Table~\ref{tab:balance_prestudy} presents balance checks on tutor characteristics, including gender, years of tutoring experience, and pre-study quality ratings.
Table~\ref{tab:attrition} reports attrition, defined as tutors who were assigned to a group but did not conduct any tutoring sessions during the study period.
After excluding tutors with no sessions, Table~\ref{tab:balance_poststudy_sessions} presents balance checks for the remaining sample, including an additional check on the amount of tutoring conducted by assignment group. Table~\ref{tab:balance_postsurvey} presents balance checks for the subsample of tutors with sessions in our sample who responded to the pre- and post-study surveys.
For all balance and attrition checks, we find no statistically significant differences between the treatment and control groups, indicating that random assignment was effective and the sample remains balanced after accounting for attrition.


\begin{table}[h!]
\centering
\resizebox{0.80\textwidth}{!}{%
\def\arraystretch{1.15}
\begin{tabular}{lccc}
\toprule
\textbf{}    & \textbf{Control} (N=444) & \textbf{Treatment} (N=427) & \textbf{\textit{p-value}} \\ 
\hline
Gender (Is Female) & 0.56 & 0.53 & 0.27 \\ 
Experience (Months) & 21 & 22 & 0.61 \\ 
Quality Rating & 0.41 & 0.42 & 0.62 \\ 
\bottomrule

\end{tabular}
}
\vspace{1em}
\caption{Balance check of pre-study sample of tutors assigned to treatment and control. Experience reports the number of months the tutor has been with the tutoring  provider. Quality Rating reports the tutor's tutoring quality rating determined prior to the study by the tutoring provider based on session observations \& quality rubric scores.\label{tab:balance_prestudy}}
\end{table}


\begin{table}[h!]
\centering
\def\sym#1{\ifmmode^{#1}\else\(^{#1}\)\fi}
\resizebox{0.6\textwidth}{!}{%
\def\arraystretch{1.15}
\begin{tabular}{l*{2}{c}}
\toprule

                    &\multicolumn{1}{c}{\shortstack{Session Attrition}}&\multicolumn{1}{c}{\shortstack{Survey Attrition}}\\
\hline
Treatment           &      -0.013         &      0.0077         \\
                    &     (0.021)         &     (0.022)         \\
[1em]
Tutor Quality Score &      -0.021         &      -0.088         \\
                    &     (0.056)         &     (0.063)         \\
[1em]
Tutor Experience (Months)&    -0.00034         &     -0.0015\sym{+}  \\
                    &   (0.00084)         &   (0.00086)         \\
[1em]
Tutor Gender: Female&       0.032         &      -0.016         \\
                    &     (0.020)         &     (0.022)         \\
\hline
N                   &         871         &         871         \\
\bottomrule 
\end{tabular}
}
\vspace{1em}
\caption{Tutor differential attrition on session participation and post-survey response.  Robust standard errors in parenthesis. Experience reports the number of months the tutor has been with the tutoring provider. Quality Rating reports the tutor's tutoring quality rating determined prior to the study by the tutoring provider based on session observations \& quality rubric scores. $^{+}p<0.1$; $^{*}p<0.05$; $^{**}p<0.01$; $^{***}p<0.001$. \label{tab:attrition}}
\end{table}


\begin{table}[h!]
\centering
\resizebox{0.95\textwidth}{!}{%
\def\arraystretch{1.15}
\begin{tabular}{lccc}
\toprule
\textbf{}    & \textbf{Control} (N=394) & \textbf{Treatment} (N=386) & \textbf{\textit{p-value}} \\ 
\hline
Gender (Is Female) & 0.55 & 0.52 & 0.33 \\ 
Experience (Months) & 21 & 22 & 0.66 \\ 
Quality Rating & 0.41 & 0.42 & 0.61 \\ 
\# Sessions during Study & 5 & 5 & 0.39 \\ 
Total Session Time during Study (Minutes) & 199 & 194 & 0.48 \\ 
\bottomrule

\end{tabular}
}
\vspace{1em}
\caption{Balance check on study's actual tutor sample with sessions (post attrition). Experience reports the number of months the tutor has been with the tutoring provider. Quality Rating reports the tutor's tutoring quality rating determined prior to the study by the tutoring provider based on session observations \& quality rubric scores. \label{tab:balance_poststudy_sessions}}
\end{table}



\begin{table}[h!]
\centering
\resizebox{0.95\textwidth}{!}{%
\def\arraystretch{1.15}
\begin{tabular}{lccc}
\toprule
\textbf{}    & \textbf{Control} (N=371) & \textbf{Treatment} (N=355) & \textbf{\textit{p-value}} \\ 
\midrule
Gender (Is Female) & 0.55 & 0.53 & 0.57 \\ 
Experience (Months) & 21 & 22 & 0.45 \\ 
Quality Rating & 0.41 & 0.42 & 0.76 \\ 
\# Sessions during Study & 5 & 5 & 0.77 \\ 
Total Session Time during Study (Minutes) & 202 & 201 & 0.91 \\ 
\bottomrule

\end{tabular}
}
\vspace{1em}
\caption{Balance check on study's tutor sample post-survey. Experience reports the number of months the tutor has been with the tutoring provider. Quality Rating reports the tutor's tutoring quality rating determined prior to the study by the tutoring provider based on session observations \& quality rubric scores. \label{tab:balance_postsurvey}}
\end{table}


\section{Tutor CoPilot Training for Treatment Tutors \label{app:tutor_copilot_training}}
Before integrating Tutor CoPilot into live tutoring sessions, we provided treatment tutors with targeted training to ensure they understood its capabilities and how to use the tool effectively. 
Proper training was critical to avoid the common pitfall in education where new technologies are introduced without adequate instruction, leading to misinterpretation, misuse, and potentially negative impacts on students.\footnote{See, for example, the ``Education and AI: Achieving equity and respecting the rights of students'' event hosted by the Brookings Institution: 
\url{https://www.brookings.edu/events/education-and-ai-achieving-equity-and-respecting-the-rights-of-students/}.} 
Without proper guidance, educators may develop incorrect expectations of the tool or use it in ways that hinder learning outcomes.

To prepare tutors, we provided the following training materials. 
We developed a slide deck illustrating various real-world tutoring scenarios that Tutor CoPilot supports and tutors went through this slide deck. 
Each scenario featured a student mistake, the lesson topic, a brief commentary on the nature of the mistake, and an expert strategy response generated by Tutor CoPilot. 
These examples are from actual tutoring interactions to make the training practical and relatable.
Afterwards, treatment tutors were paired in a buddy system with another treatment tutor who would also get access to Tutor CoPilot. 
With their pair, they role-played in the ``student'' and ``tutor'' role. 
When a tutor played the ``tutor'' role, they had access to Tutor CoPilot and could see how Tutor CoPilot responded in real-time. 
The entire training process took approximately 2-3 weeks for all treatment tutors to complete.

\section{Study data\label{app:study_data}}

\paragraph{Tutors.} All information about tutors was obtained from the tutoring provider. We received information on their gender (Male or Female), their pre-study quality rating (continuous between -1 and 1), and their pre-study tutoring experience counted as the number of months they've worked with the tutoring provider.
The tutor's quality rating is determined by the tutoring provider based on human observations of the tutor's tutoring sessions and scored on the provider's proprietary rubric. 
These scores are averaged to produce the quality rating used in our study. 
We did not receive information on the tutor's experience prior to their employment with our partner tutoring provider for this study.
We performed tutor-level randomization and confirmed that these covariates are balanced between groups.
We report the balance checks in Supplementary Information section~\ref{app:balance}.

\paragraph{Students. } For students, we received information on their gender (Male or Female), Race/Ethnicity (Hispanic, White, Black, Asian, Pacific Islander, American Indian or Alaska Native, Two or more races), whether they receive Free and Reduced lunch (Yes or No), whether they are in the Limited English Proficiency program (Yes or No) and their pre-study baseline and post-study NWEA MAP Math and Reading Scores. 
The NWEA MAP is a standardized test administered three times a year, tracking students' academic growth over time.
For more information on NWEA MAP, please refer to \url{https://www.nwea.org/map-growth/}.

\paragraph{Sessions.} 
We receive all the session-level data from our two-month study, which started at the end of March. 
We excluded sessions that were conducted with part-time tutors, as classified by the tutoring provider.
The session-level data included the session's chat transcript, whiteboard activity, post-session survey responses from the student and tutor, the tutor's treatment assignment, and Tutor CoPilot use if applicable.
The Tutor CoPilot use includes information on the session id, the request timestamp, conversation context preceding the click, the expert strategy, and corresponding suggestion generated by Tutor CoPilot. 
Our study includes 2,000+ uses of Tutor CoPilot.

\section{Development of NLP Classifiers \label{app:nlp_classifiers}}

To analyze the use and impact of Tutor CoPilot, we want to identify the tutoring moments in which the tool was used and the pedagogical strategies used by tutors, as measured in their language. 
However, identifying these features requires overcoming several technical challenges.
First, the chat transcripts, which span over 350,000 messages, are \textit{unlabeled} for the moments and strategies. 
Therefore, we need a scalable and consistent method to label the data efficiently. 
Second, some categories, such as strategies that we consider high-quality, may be underrepresented in the data. 
Therefore, we need methods that can handle long-tailed distributions to handle rare categories. 
Finally, some tutoring contexts may involve actions that are not directly observable in the chat transcripts, such as students working on a whiteboard, leading to incomplete data. 
Therefore, the final method must also account for these contextual gaps.
To address these challenges, we developed novel natural language processing (NLP) methods capable of efficiently labeling our entire dataset. 
Our approach involved three key steps: 

\begin{enumerate}
    \item \textbf{Taxonomy Development with Unsupervised Methods:} We began by creating a novel taxonomy of tutoring moments and pedagogical strategies. We constructed the taxonomies leveraging unsupervised methods (e.g., topic modeling) to efficiently organize the data and inform the development of the taxonomy. These taxonomies serve as the foundation for our analysis.
    \item \textbf{Classifier Dataset Construction and Training for Handling Imbalanced, Imperfect Data:} Next, we label a dataset of 3,000 examples, annotated according to our taxonomies; we will refer to this dataset as our \textit{classifier dataset}. We train and evaluate the classifiers on this data, leveraging the imbalanced nature of categories to automatically reweigh our data and contextual information to mitigate the effects of incomplete data.
    \item \textbf{Application of Classifiers on Downstream Datasets:} We run inference with the classifiers on our \textit{downstream datasets}. We apply the moments classifier on our Tutor CoPilot usage data, which includes the conversation context leading up to the click of Tutor CoPilot. We apply the strategies classifier to both treatment and control chat sessions; we measure the causal impact of Tutor CoPilot on language by comparing the prevalence of strategies between groups.
\end{enumerate}

The process of developing the taxonomy and training the classifiers is consistent across both moments and strategies; this process was agnostic to the treatment assignment condition. 
Note that the application of these classifiers differs based on the downstream datasets they are applied to. We will now elaborate on these steps in more detail. 

\subsection{Taxonomy Development}
We initially developed a taxonomy for tutoring moments and strategies based on prior research and language patterns observed in tutors and students on the platform.
To refine this taxonomy, we used the BERTopic package~\citep{grootendorst2022bertopic}, which allowed us to organize the diverse language used and clearly define the boundaries between categories.
This process ensured that our taxonomy accurately captured the key distinctions needed for effective analysis.
For both taxonomies, we used \texttt{all-MiniLM-L6-v2} as our pretrained embedding model, and a count vectorizer on bigrams and trigrams as our vectorizer model; from qualitative experiences with tutoring language, we omitted unigrams because they tend not to cluster the text in pedagogically interesting ways.

\paragraph{Taxonomy of tutoring moments.} 
We structured the taxonomy around the flow of tutoring sessions, beginning with typical phases: starting the session, working on practice problems, and completing exit tickets. 
Because we are particularly interested in moments of learning and tutor support---such as distinguishing moments when the tutor introduces the problems from moments when the tutor supports the student’s attempt at the problem---we refined the moments into ``start,'' ``during student attempt,'' and ``after student attempt'' to capture these distinctions.

We further refined the definitions and boundaries of these moments by running BERTopic across the entire dataset. 
While this unsupervised approach highlighted useful patterns, it struggled with less common language, context-sensitive utterances, or unique ways that tutors engaged with students. 
Because we cared about capturing these patterns as well, we took a supervised learning approach to classify the tutoring moments more effectively but used the emergent topics to inform the taxonomy.
Our final taxonomy of tutoring moments is shown in Table~\ref{tab:moments}.

\paragraph{Taxonomy of strategies.}
We constructed the taxonomy of strategies first by using the strategies preferred and dispreferred by experts from prior work Wang et al.~(2024a)~\citep{wang-etal-2024-bridging}. 
The strategies preferred by experts were ``explain a concept'', ``ask a question'', ``provide a hint'', ``provide a problem-solving strategy'', ``provide a worked example'', ``provide a minor correction'', ``provide a similar problem'', ``simplify the question'', ``affirm the correct answer'' and ``encourage the student''. 
The strategies dis-preferred by experts were ``give away the answer/explanation'' to that specific problem or just asking the student to ``recheck/retry'' without any further guidance. 

Again, we used topic modeling to refine the boundaries of these categories. 
For example, we realized that the ``ask a question'' category was too broad. Questions vary a lot in their pedagogical quality and usefulness. 
Questions can be used to merely check for understanding, like ``Did you understand my explanation?'' and questions can also be used to guide the student’s thinking and be more process-oriented, such as ``Which place value should be compared?'' 
Based on these insights complemented by prior literature, we refined the ``ask question'' category to focus on questions that actively guided the student’s thought process. 

Not every message can be categorized into these strategies. Thus, we also had an N/A category, to capture things that should be ignored. This category included: 
\begin{itemize}
    \item Transition language: e.g., Let's do the next problem. or It's time to show your mastery in today's session.
    \item Checking in with the student: e.g., Are you there? or Let's focus on the session.
    \item Starting or ending the session: e.g., Let's start the session.
    \item Rushing the student: e.g., We are running out of time.
    \item Small talk: e.g., How was your day? 
    \item Talk related to points: e.g., You receive an additional point for your efforts.
    \item Talk related to the platform: e.g., Let me increase my pace.
    \item Question instructions: e.g., In this question, you need to find the ordered pair represents a vertex of the parallelogram that is a reflection of Point S across the y-axis.
    \item Prompts: e.g., “Let me know if you need any help along the way.” or “Give it a try.”
\end{itemize}

\subsection{Classifier Dataset Construction and Training on Imbalanced, Imperfect Data}

To annotate our downstream dataset, we need to train classifiers on a subset of the data for the classifiers to reliably label for the moments and strategies. 
First, we subsampled a dataset of 3,000 examples that had a balanced representation from both treatment and control chat transcripts. Each example consisted of a pair of the context (10 prior messages) and the tutor’s message following this context. 

To efficiently label the classifier dataset, we adopted a Human+LLM annotation approach. 
This method follows recent trends in combining human expertise with AI to scale annotation efforts~\citep{wang2024human, kim2024meganno+}. 
We prompted an LLM to first annotate the dataset using our taxonomies, after which human annotators (two co-authors) reviewed and corrected the labels. 
Unlike other studies on Human+LLM annotation approaches that start with an existing codebook, our codebook combined unsupervised methods to inform the taxonomy developed as previously described. 

We split the annotated dataset for both moments and strategies into training, validation, and test sets in a 6:1:3 ratio. 
We cast this setting as a multiple binary classification task; this provides multiple advantages, such as this doesn’t assume mutual exclusiveness among classes, which aligns well with real-world data where few classes might be similar with each other, and each class is considered independent with its own predictor. 
This is a nice property since real-world data often has more than one semantic label. 

We finetuned a RoBERTa large model and introduced new tokens to separate the context and target text to be classified: \texttt{[CONTEXT\_TOKEN] \{context\} [TARGET\_TOKEN] \{utterance\}}. 
Given the long-tailed nature of some categories, we automatically reweighed the data based on class distributions and optimized the classifier using a sigmoid cross-entropy class-balanced loss~\citep{cui2019class}.
The class-balanced loss introduces a weighting factor that is inversely proportional to the effective number of samples needed to train a good classifier. 
We ran a hyper-parameter sweep over the the loss' hyper-parameters and learning rate on the validation set, selecting the hyper-parameters that yielded the lowest validation loss. 
The performance of the classifiers is reported using the F1 score on the test set, shown in the taxonomy tables in the main paper.

We set a priori thresholds on the test F1 scores of 0.60+ for categories to include in our downstream analysis. 
Categories that did not meet this threshold were excluded from further analysis. 
This led us to dropping ``during exit ticket attempt'' from the moments analysis, and ``explanation of concept'' and ``provide hint'' from the strategies analysis.

\subsection{Application of Classifiers on Downstream Datasets}

After training and validating our classifiers, we applied them to our downstream datasets to answer our research questions.

\paragraph{Moments.} 
To understand the tutoring moments in which treatment tutors used Tutor CoPilot, we run inference with our moment classifiers on the Tutor CoPilot usage data. 
We use this dataset to identify the conversation context leading up to the tutor’s click on the tool and the message that followed. We formatted this data the same as the training format. 
After running inference on the dataset, we report the frequency of these identified moments within the Tutor CoPilot usage data. 
By doing so, we tied these frequency findings back to our research question on the type of tutoring moments tutors used Tutor CoPilot. 

\paragraph{Strategies.} 
To measure the the causal language impact of Tutor CoPilot, we run inference with our strategy classifiers on all chat transcripts. The resulting dataset includes 241,005 classified tutor messages.
We then separate the transcripts into treatment and control groups based on the tutor’s treatment assignment, and identify which strategies are more prevalent in one group over the other. 
To quantify these differences, we use a logit regression specification with standard errors clustered by student-tutor pairs to account for repeated interactions. We report the treatment coefficient for each strategy in Figure~\ref{fig:strategies_clustered}, with additional statistics and the calculated odds ratios in Table~\ref{tab:strategies_clustered}.

\paragraph{Overlap in Strategy Classifiers} Table~\ref{tab:strategy_overlap} shows the overlap in strategy classifications for each message in our sample as a proportion of the total number of messages.

\begin{table}[h!]
\centering
\resizebox{\textwidth}{!}{%
\def\arraystretch{1.15}
\begin{tabular}{lccccccc}
\toprule
 & \shortstack{Ask \\ Question \\ to Guide \\ Thinking} & \shortstack{ Give \\ Solution \\ Strategy} & \shortstack{ Prompt \\ Student \\ to Explain} & \shortstack{Encourage \\ Student in \\ Generic \\ Way} & \shortstack{Affirm \\ Student's \\ Correct \\ Attempt} & \shortstack{Give \\ Away \\ Answer/ \\ Explanation} & \shortstack{Ask \\ Student \\ to Retry} \\
\midrule
Ask Question to Guide Thinking & 0.04451 & 0.00005 & 0.00000 & 0.00007 & 0.00027 & 0.00076 & 0.00010 \\
Give Solution Strategy & 0.00005 & 0.00509 & 0.00000 & 0.00000 & 0.00005 & 0.00047 & 0.00004 \\
Prompt Student to Explain & 0.00000 & 0.00000 & 0.02545 & 0.00008 & 0.00023 & 0.00000 & 0.00017 \\
Encourage Student in Generic Way & 0.00007 & 0.00000 & 0.00008 & 0.11845 & 0.02031 & 0.00003 & 0.00033 \\
Affirm Student's Correct Attempt & 0.00027 & 0.00005 & 0.00023 & 0.02031 & 0.13405 & 0.01412 & 0.00034 \\
Give Away Answer/Explanation & 0.00076 & 0.00047 & 0.00000 & 0.00003 & 0.01412 & 0.05917 & 0.00008 \\
Ask Student to Retry & 0.00010 & 0.00004 & 0.00017 & 0.00033 & 0.00034 & 0.00008 & 0.00935 \\
\bottomrule
\end{tabular}
}
\vspace{.5em}
\caption{Overlap in strategy classification results. Each value represents the proportion of messages with a positive result by the row and column strategy classifiers. The diagonal presents the proportion of messages classified as one specific strategy.\label{tab:strategy_overlap}}
\end{table}


\section{Study Costs \label{app:study_cost}}
The total API cost for 429 treatment tutors over the 2-month study was \$1,419.66, resulting in an estimated annual cost of \$20 per tutor.

\renewcommand{\bibsection}{\section*{References for Supplementary Information}}


\end{bibunit}           


\end{document}